\documentclass{article} % onecolumn (standard format)

\usepackage[utf8]{inputenc}
\usepackage{amsmath}
\usepackage{amsfonts}
\usepackage{amssymb}
\usepackage{listings}
\usepackage{mathrsfs}
\usepackage{times}
\usepackage{float}
\usepackage{color}
\usepackage{setspace}
\usepackage{color,soul}

\usepackage{amsmath,amsfonts,amsthm,eucal}
\usepackage{enumerate}
\usepackage[letterpaper,margin=3cm]{geometry}
\usepackage{float}
\usepackage[title]{appendix}
\usepackage{color}
\usepackage{tabularx}
\usepackage{siunitx}

\usepackage[
pdfstartview=XYZ,
bookmarks=true,
colorlinks=true,
linkcolor=blue,
urlcolor=blue,
citecolor=blue,
pdftex,
bookmarks=true,
linktocpage=true,   % makes the page number as hyperlink in table of content
hyperindex=true
]{hyperref}

\usepackage{lipsum}
\usepackage[FIGBOTCAP,TABTOPCAP,bf,tight]{subfigure}

\usepackage{natbib}
\usepackage[pdftex]{graphicx}
\usepackage{epstopdf}
\usepackage{booktabs} 

\usepackage{tikz,mathpazo}
\usetikzlibrary{shapes.geometric, arrows}

\usepackage{caption}
\newcommand{\tensor}[1]{\ensuremath{\boldsymbol{#1}}}

\theoremstyle{remark}

\renewcommand{\vec}[1]{\ensuremath{\boldsymbol{#1}}}

\usepackage{algorithm}
\usepackage[noend]{algpseudocode}

\title{Meta-modeling game for deriving theoretical-consistent, micro-structural-based traction-separation laws via deep reinforcement learning} 

\begin{document}

%\titlerunning{Automated generation of traction-separation laws}

\author{Kun Wang       \and
        WaiChing Sun 
         \thanks{Corresponding author: WaiChing Sun, 
Assistant Professor, Department of Civil Engineering and Engineering Mechanics, 
 Columbia University , 
 614 SW Mudd, Mail Code: 4709, 
 New York, NY 10027
  \textit{wsun@columbia.edu}  }}

%\date{Received: \today / Accepted: date}
\maketitle

\begin{abstract}
This paper presents a new meta-modeling framework to employ deep reinforcement learning (DRL) to generate mechanical constitutive 
models for interfaces. 
The constitutive models are conceptualized as information flow in directed graphs. 
The process of writing constitutive models are simplified as a sequence of forming graph edges with the goal 
of maximizing the model score (a function of accuracy, robustness and forward prediction quality). 
Thus meta-modeling can be formulated as a Markov decision process with well-defined states, actions, rules, objective functions and rewards. 
By using neural networks to estimate policies and state values, 
the computer agent is able to efficiently self-improve the constitutive model it generated through self-playing, in the same way AlphaGo Zero (the algorithm that outplayed the world champion in the game of Go)
improves its gameplay. 
Our numerical examples show that this automated meta-modeling framework not only produces models which outperform existing cohesive models on benchmark traction-separation data, 
but is also capable of detecting hidden mechanisms among micro-structural features and incorporating them in constitutive models to improve the forward prediction accuracy, which are difficult tasks
to do manually. 
\end{abstract}
%\keywords{meta-modeling; traction-separation law; data-driven computational mechanics; path-dependent responses; fracture opening and closure}

\section{Introduction}
\label{intro}
Constitutive responses of interfaces are important for a wide spectrum of problems that involve spatial domain with embedded strong discontinuity, such as fracture surfaces \citep{rice1968path, park2009unified, wang2017unified, BRYANT2018}, slip lines  
\citep{rabczuk2006new, borja2007continuum}, joints \citep{elices2002cohesive}
and faults \citep{ohnaka1989cohesive, wang2018multiscale, sun2018prediction}. 
While earlier modeling efforts, in particular those involving the modeling of cohesive zones 
, often solely focus on mode I kinematics, the mixed mode predictions of traction-separation 
law relations are critical for numerous applications, ranging from predicting damage upon
impacts \citep{ortiz1999finite}, to predicting seismic events \citep{rudnicki1980fracture}. 
Recent work by \citet{park2009unified} provide a comprehensive account of the major 
characteristics of traction-separation laws and conclude that, while there are differences in details, 
most of the traction-separation laws obey a number of universal principles, such as the indifference 
of any superimposed rigid-body motion, the finite work required to create new surface, the existence 
of characteristics length scales, and the vanishing of cohesive traction with sufficient separations. 

In the case where the loading history is not monotonic, constitutive responses of interfaces often 
become path-dependent. For instance, geomaterials, such as fault gauges, are known to exhibit 
rate- and state-dependent frictional responses 
\citep{paterson2005experimental, sun2013unified, borja2013plasticity, wang2016identifying, na2017computational}. 
While there are phenomenological models designed to capture the path-dependent responses of the interfaces, 
a recent trend that gains increasing popularity is to replace the phenomenological traction-separation laws
with computational homogenization procedure to capture the responses of materials with heterogeneous
microstructures (cf. \citet{moes2003computational, hirschberger2008computational, hirschberger2009computational}. 
Nevertheless, as pointed out previously in \citet{wang2018multiscale}, the major issue of applying hierarchical multiscale 
coupling on interfacial problems is the increasing computational demand due to the large number of required representative elementary simulations, a trade-off that is widely known in FEM$^{2}$ \citep{feyel2003multilevel} and other homogenization-based multiscale methods, such as FEM-DEM \citep{sun2011connecting, sun2011multiscale,  fish2013practical, sun2013multiscale, wang2015anisotropy, liu2016nonlocal,  kuhn2015stress, wang2016semi, wu2018multiscale}.

To overcome this computational barrier, surrogate models are often derived to replicate the homogenized responses of sub-scale simulations
\citep{kirane2008cold, verhoosel2010computational, keshavarz2013multi, panchal2013key, faisal2014computational, liu2016determining, tallman2017reconciled}.
Nevertheless, since surrogate models are often constitutive laws hand-crafted by modelers to incorporate morphology-dependent features
\citep{liu2016determining}, deriving, verifying and validating a surrogate model that can incorporate the essential information to yield macroscopic predictions with sufficient accuracy and robustness remain difficult and time-consuming. 
Data-driven models such as \citet{le2015computational, bessa2017framework, versino2017data, kafka2018data} and \citet{wulfinghoff2018model} attempted 
to overcome this issue via supervised machine learning (e.g. neural network \citep{lefik2002artificial}, symbolic regression model \citep{versino2017data}) and unsupervised machine learning (e.g. dimensional reduction, feature extraction and clustering \citep{bessa2017framework, wulfinghoff2018model}). 

In particular, recent work by \citet{wang2018multiscale} attempted to resolve this issue by building a generic recurrent neural network that can 
easily incorporate different types of sub-scale information (e.g. porosity, fabric tensor, and relative displacement) to predict traction. This technique uses the concept of directed graph on the transfer learning approach (cf. \citet{pan2010survey}) in which multiple neural networks trained to make predictions on other physical quantities (e.g. relationship between porosity and fabric tensor) are re-used to generate additional inputs for predicting traction. 
However, the determination of the optimal input information (in addition to the displacement jump history) and configurations of information flow that enhances the prediction accuracy still requires a time-consuming trial-and-error task (cf. Section 4.3 \citet{wang2018multiscale}). 

In this work, we introduce a general artificial intelligence approach to automate the creation and validation
of traction-separation models. Unlike the previous approach in which neural networks are often
used to either identify material parameters or create black-box constitutive laws, this work focuses on 
leveraging the capacity of a computer to improve via self-playing, a technique commonly 
referred as (deep) reinforcement learning in the computer science community 
\citep{sutton1992introduction, silver2016mastering, silver2017mastering}. 
In the past two years, the functionality of algorithms automatically generated from deep reinforcement learning have 
achieved remarkable success. In many cases, the  demonstrated capacities were 
thought to be impossible in the past.  
For instance, the algorithm trained by deep reinforcement learning created by a company called DeepMind is able to 
outperform human experts in Go, Chess and Atari games. 
The most exciting part of this achievement is that, unlike previous AI such as the IBM Deep Blue, 
the deep reinforcement learning does not rely on hand-crafted policy evaluation functions and is therefore applicable to different kinds of games once they are defined and implemented. 

This success motivates this research of proposing a meta-modeling approach where deep 
reinforcement learning may generate constitutive laws for (1) a given set of data, (2) a well-defined objective, 
and (3) a given set of universal principles. 
To achieve this goal, we recast the process of writing a constitutive model as a game with components suitable for deep reinforcement learning, involving a sequence of actions completely compatible with the stated rules (i.e., the law of physics). 
First, we define the model score, which could be any objective function suitable for a given task. 
For instance, this objective can be minimizing the discrepancy between calibrated experimental results and \textbf{\textit{blind predictions}} measured by a norm, or a constrained 
optimization problem that gives considerations on other attributes such as consistency, speed, and robustness \citep{wang2016identifying}. 
Once the score (i.e., the objective) is clearly defined, we then implement the rules, which are the universal principles of mechanics, such as material frame indifference, laws of thermodynamics. 
These rules are applied in an environment in which scores are sampled. 
In the case of traction-separation law, the environment is simply the validation process itself. 

Following this, we then define the action space which consists of 
a number of actions available for the modelers to write constitutive models. 
Once the action space and the model score are defined, we leverage the directed graph modeling technique to generate a state. 
The state at the end of each game represents a constitutive model automatically generated from the computer algorithm. 
In reality, the action space could be of very high dimensions such that manually deriving, implementing, verifying and validating all possible configurations are not feasible. 
This situation is similar of playing the games of chess and Go where the number of possible combinations of decisions or moves (each can be represented by a decision tree) remains finite but is so enormous that it is not possible to seek the optimal moves by exhausting all possibilities \citep{shannon1950xxii}. 
%However, we will simplify the problem by considering only the common actions used in previous literature and for traction-separation laws surveyed in \citet{park2009unified} and the deep neural network model published in \citet{wang2018multiscale} to demonstrate the idea. 

With the state, action, rule and objective defined, the most critical part is to assign reward for each action. 
In principle, if the action space is of very low dimension, i.e., there are not many ways to model the physical processes, 
then the reward for each action can be determined by exhausting all the possible model configurations. 
However, in the case of writing a complex traction-separation model, we cannot evaluate the quality of the model until its predictions are compared with benchmark data. 
Therefore, the ability to approximate the reward for each action (in our case the modeling choices)
without the need to evaluate all the available options becomes crucial for the success of the meta-modeling approach. 

The deep reinforcement learning is therefore ideal for us to achieve this goal. 
We can approximate the rewards via neural networks and the Bellman expectation equation \citep{bellman1957markovian, dolcetta1984approximate}. 
By repeatedly generating new constitutive laws (i.e., playing the game of writing models), the agent will use the reward obtained
from each played game, in analogy with the binary game result (win/loss) at the end of a Go game, 
to update the action probabilities and value functions 
to improve the agent's ability to write good constitutive laws. 
Through sufficient self-plays, the reinforcement learning algorithm
then improves the modeling choices it made over time until it is ready for predictions.  

\begin{figure}[h!]\center
		\includegraphics[width=0.95\textwidth]{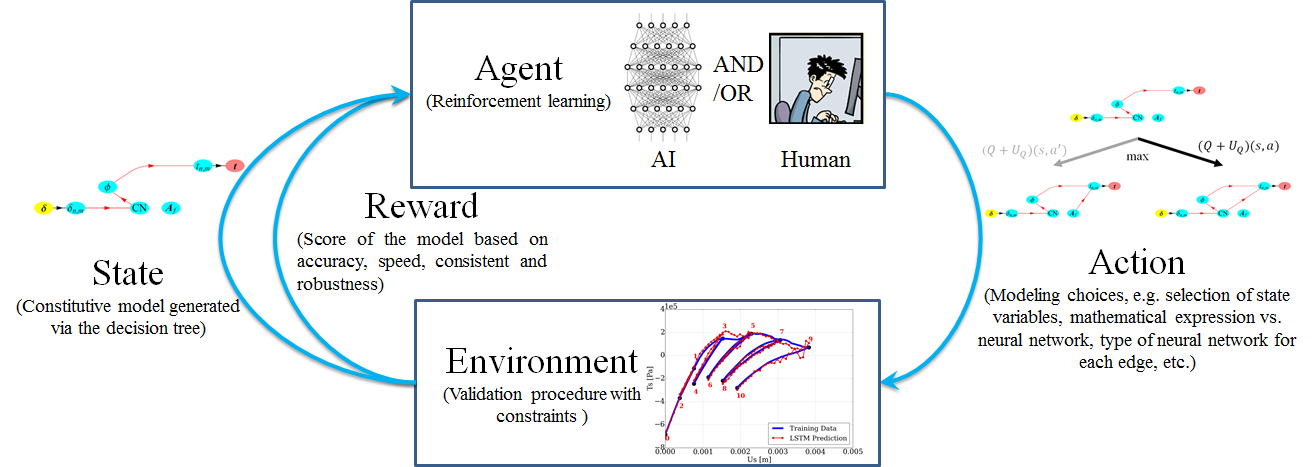}
	\caption{Scheme of the reinforcement learning algorithm in which an agent interacts with environment and receives rewards. Through exploration, the agent then determines better actions to achieve a particular goal defined by the reward. In our case, the reward is the score which represents the quality of the forward prediction, the action is any possible activities required to derive a constitutive law, and the environment is the procedure that compares the predictions with the benchmark data.}
	\label{fig:agent}
\end{figure}

There are a few major upshots for this approach. 
First, once the reinforcement learning algorithm is established, it can serve as a model generator without any human intervention. 
Second, since we regard the validation process as the environment component of the reinforcement learning, the performance of a resultant model is simultaneously evaluated and therefore validations are always a part of the model writing process. 
Third, the meta-modeling approach may easily embed any existing model generated by domain experts into the action space without re-implementing a new model. 
These unique capabilities enable us to have an unbiased tool to evaluate how well existing models fulfill a particular objective. 
Furthermore, since the model generation procedure is automated once an objective function is defined, this work may potentially eliminate the need of writing multiple incremental models for the same materials over time. 
Finally, this modeling approach is particularly powerful for discovering hidden physical coupling mechanisms that are otherwise too subtle to detect with human observation. 

The rest of the paper is organized as follows. 
We first review the directed graph approach that enables us to generate and utilize a decision tree to represent the modeling process (Section \ref{sec:graph}). 
The definition of model scores is then described in Section \ref{sec:objective_score}. 
We then provide a formal definition of a game invented to generate traction-separation laws for predictions (Section \ref{sec:game}). 
This is followed by a description on how to use the reinforcement learning for the traction-separation law generation (Section \ref{sec:gameplay}). 
Two numerical experiments are then used to showcase the performance of the automated meta-modeling approach using synthetic data from microscale discrete element simulations (Section \ref{sec:numexperiment}). 
The major findings are then summarized in the conclusions. 

\section{Representing traction-separation law in directed graph}  \label{sec:graph}
%Kun --- can you make some drawing in Mathematica to show ---like 9 to 10 different directed graphs for cohesive zone model? I will talk to you tomorrow. 
In this section, we introduce a building block for a simplified and extensible game that generates traction-separation laws by considering the relationships among different types of data collected from sub-scale simulations. 
In this game,  the goal is to find a specific way to link different types of data such that  a score function is maximized. 
Before we introduce the formal definition of the game, one necessary step is to recast the algorithm that leads to predictions from constitutive laws as a network of unidirectional information flow, i.e., a directed graph (also referred as digraph) \citep{sun2013stabilized, sun2014modeling, sun2015stabilized, salinger2016albany, wang2018multiscale}. 
Recall that a digraph $D = (V,E)$ is an ordered pair of non-empty finite sets which consists of a vertex set V and an edge set E \citep{bang2008digraphs}. 
Each edge connects a source vertex (tail) to a target vertex (head). 
Following the treatment in \citet{sun2015stabilized} and \citet{wang2018multiscale}, the following \textbf{rules} are applied to generate the traction-separation law. 
\begin{enumerate}
	\item The traction $\vec{t}$ is placed as the only leaf of the digraph (i.e., the vertex that is not source to any other vertices).
	\item The displacement jump $\vec{\delta}$ is placed as the only root of the digraph (i.e., the vertex that is not target of any other vertices). 
	\item There may exist isolated vertices in the digraph, i.e., some internal variables or microstructural features between $\vec{\delta}$ and $\vec{t}$ may not contribute to the final completed digraph and the corresponding constitutive model.
	\item The digraph is acyclic, which means that there must be no cycle in the digraph.
	\item If a vertex has sources or targets connected to it, it must be on at least one of the paths leading from $\vec{\delta}$ to $\vec{t}$. 
	This ensures that an internal variable, once considered, is fully incorporated into the final constitutive model.
\end{enumerate}

In the previous published work (cf. \citet{sun2015stabilized, wang2018multiscale}), we prescribed theoretical models or, in some cases, neural network models to create linkages and enforce the hierarchy among physical quantities (e.g. porosity-permeability relation).
While this treatment is convenient for software engineering and code design \citep{salinger2016albany}, 
this approach only works if we have a prior knowledge about the relationships among the physical quantities. 
While one may presumably make ad hoc assumptions to
complete the models, such a treatment is often at the expense of robustness. 
Another possible remedy is to gather all the measurement and data one may possibly obtain from observations and experiments, then find 
the key mechanisms that incorporate the most essential physics (e.g. the critical state plasticity for soil).
This latter approach can be re-expressed as a problem in the directed graph in which we only know the elements of the vertex set 
but have no idea whether and how these vertices are connected, except that the traction is the leaf and the displacement jump is the root of the 
directed graph. 
Note that, in reality, the creation of a deterministic constitutive law does not only limit at determining connections among vertices (physical quantities), but also includes finding hidden vertices and appropriate edges. 
These actions are not modeled in this paper, but will be considered in future studies. 
Furthermore, while our focus in this paper is on deriving the traction-separation laws, in principle, the idea can be easily extended to other problems, such as the stress-strain relation for bulk materials, the porosity-temperature-fabric-tensor-permeability relations for porous media, among others. 

For demonstration purposes, we consider a constitutive law $\vec{t}(\vec{\delta}, \vec{q})$ that predicts 
the traction vector $\vec{t}$ based on the history of the displacement jump $\vec{\delta}$ over a cohesive or cohesive-frictional surface with the normal direction vector being $\vec{n}$. 
$\vec{q}$ is a collection of state variables with $n$ degrees of freedom, i.e., $q_1, q_2, q_3, ..., q_n$. 
We use sub-scale discrete element simulations to generate synthetic data and attempt to create a traction-separation model which 
can replicate the constitutive responses of complex loading histories. 

Imposing restrictions of material frame indifference and assuming isotropic cohesive-frictional surface, the traction-separation model can be simplified to \citep{ortiz1999finite}
\begin{equation}
\vec{t}(\vec{\delta}, \vec{q}) = \vec{t}(\delta_n, \delta_m, \vec{q}),
\end{equation}
where $\delta_n = \vec{\delta} \cdot \vec{n}$ and $\delta_m = |\vec{\delta_m}| = |\vec{\delta} - \delta_n \vec{n}|$. 
Hence, the traction $\vec{t}$ is related to its components $t_n$ and $t_m$ that
\begin{equation}
\vec{t}(\vec{\delta}, \vec{q}) = t_n(\delta_n, \delta_m, \vec{q}) \vec{n} + t_m(\delta_n, \delta_m, \vec{q}) \frac{\vec{\delta_m}}{\delta_m}.
\end{equation}

%A common simplification of the formulation of mixed-mode cohesive laws (\citep{camacho1996computational, ortiz1999finite, park2011cohesive}) introduces an effective displacement jump 
%$\delta = \sqrt{\delta_n^2 + \beta^2 \delta_m^2}$ and an effective traction $t=\sqrt{t_n^2 + \beta^{-2} t_m^2}$. $\beta$ is a parameter assigning different weights to the normal and sliding displacement jumps.
%Hence an alternative form is written only in terms of effective displacement jumps and tractions as
%\begin{equation}
%\vec{t}(\vec{\delta}, \vec{q}) \rightarrow t(\delta, \vec{q}).
%\end{equation}

The internal variables in $\vec{q}$, if the cohesive surface is composed of a thin layer of granular materials, can be chosen among a large set of geometrical measures on micro-structural attributes \citep{sun2013multiscale, kuhn2015stress}. 
In this work, we first manually select the following measures to be the intermediate vertices (the vertices that are neither the leaves nor the roots) to make forward predictions on the traction vector. 
\begin{itemize}
	\item Porosity $\phi$, the ratio between the volume of the void and the total volume of a representative volume element (RVE) of the material layer. 
	\item Coordination number $CN = N_{\text{contact}}/N_{\text{particle}}$ where $N_{\text{contact}}$ is the number of particle contacts and $N_{\text{particle}}$ is the number of particles in the RVE. 
	\item Fabric tensor $\tensor{A}_{f} = \frac{1}{N_{\text{contact}}}\sum_{c=1}^{N_{\text{contact}}} \vec{n}^{c} \otimes \vec{n}^{c}$, where $\vec{n}^{c}$ is the normal vector of a particle contact $c$, $c$ = 1, 2, ...,$N_{\text{contact}}$ in the RVE.
	\item Strong fabric tensor $\tensor{A}_{sf} = \frac{1}{N_{\text{strongcontact}}}\sum_{c=1}^{N_{\text{strongcontact}}} \vec{n}^{c} \otimes \vec{n}^{c}$, where $\vec{n}^{c}$ is the normal vector of a strong particle contact (having a compressive normal force greater than mean contact force) $c$, $c$ = 1, 2, ...,$N_{\text{strongcontact}}$ in the RVE. 
\end{itemize}

All particle contacts inside the RVE can form a graph with particles as vertices and interactions as edges. 
Some quantitative measures of this graph of connectivity can be included in the internal variables $\vec{q}$ as additional microstructural characteristics. 
Here, we focus on four measures, which are computed using the software package NetworkX (\citet{hagberg2008exploring}), and their detailed explanations can be found in the software documentation. 
\begin{itemize}
	\item $d_{a}$, degree assortativity, a scalar value between -1 and 1 measuring the similarity of connections in the graph with respect to the node degree.
	\item $c_{t}$, transitivity coefficient, $c_{t}=3\frac{n_{triangles}}{n_{triads}}$, the fraction between the number of triangles and the number of triads present in contact graph.
	\item $l_{sp}$, average shortest path length in the contact graph.
	\item $\rho_{g}$, density of the graph, $\rho_{g}=\frac{2m}{n(n-1)}$, where $n$ is the total number of nodes and $m$ is the total number of edges in the graph.
\end{itemize}

\begin{figure}[h!]\center
	\subfigure[The digraph is admissible.]{
		\includegraphics[width=0.45\textwidth]{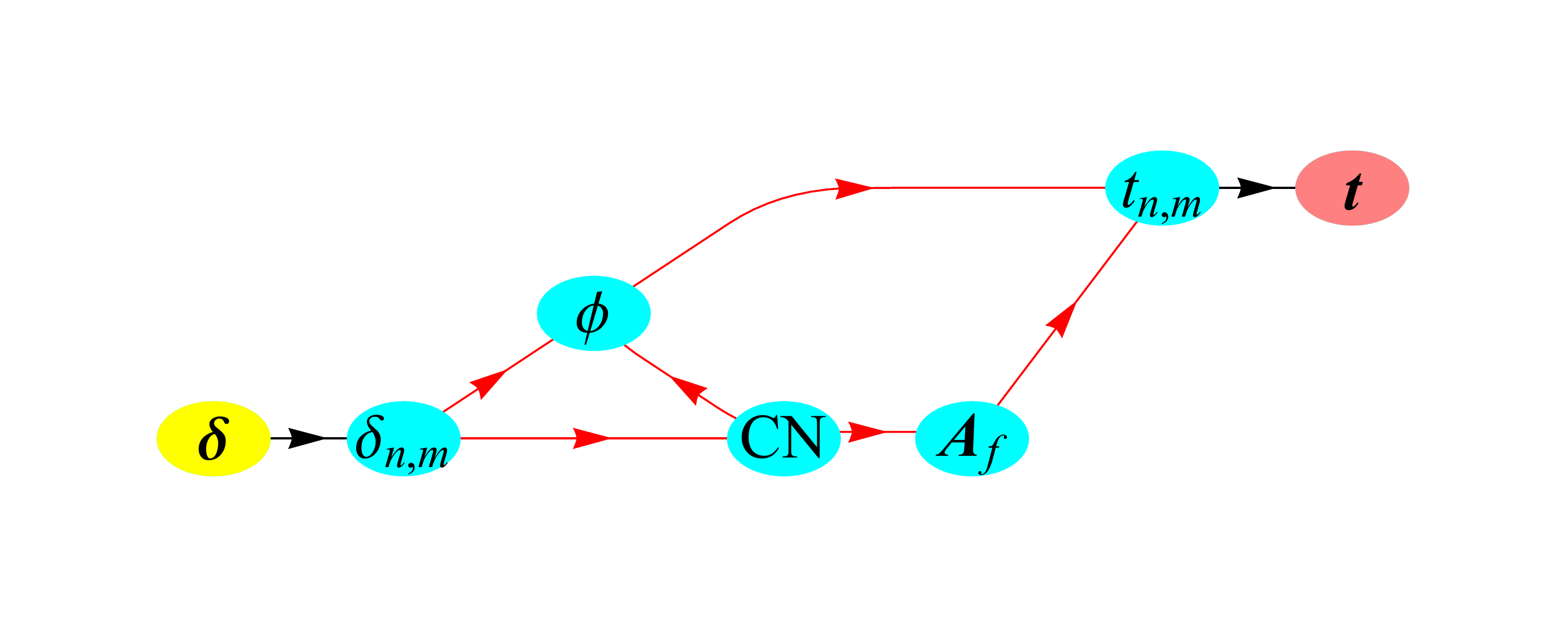}
	}
	\subfigure[The digraph is admissible according to Rule No. 3, $CN$ is not considered in the constitutive model.]{
		\includegraphics[width=0.45\textwidth]{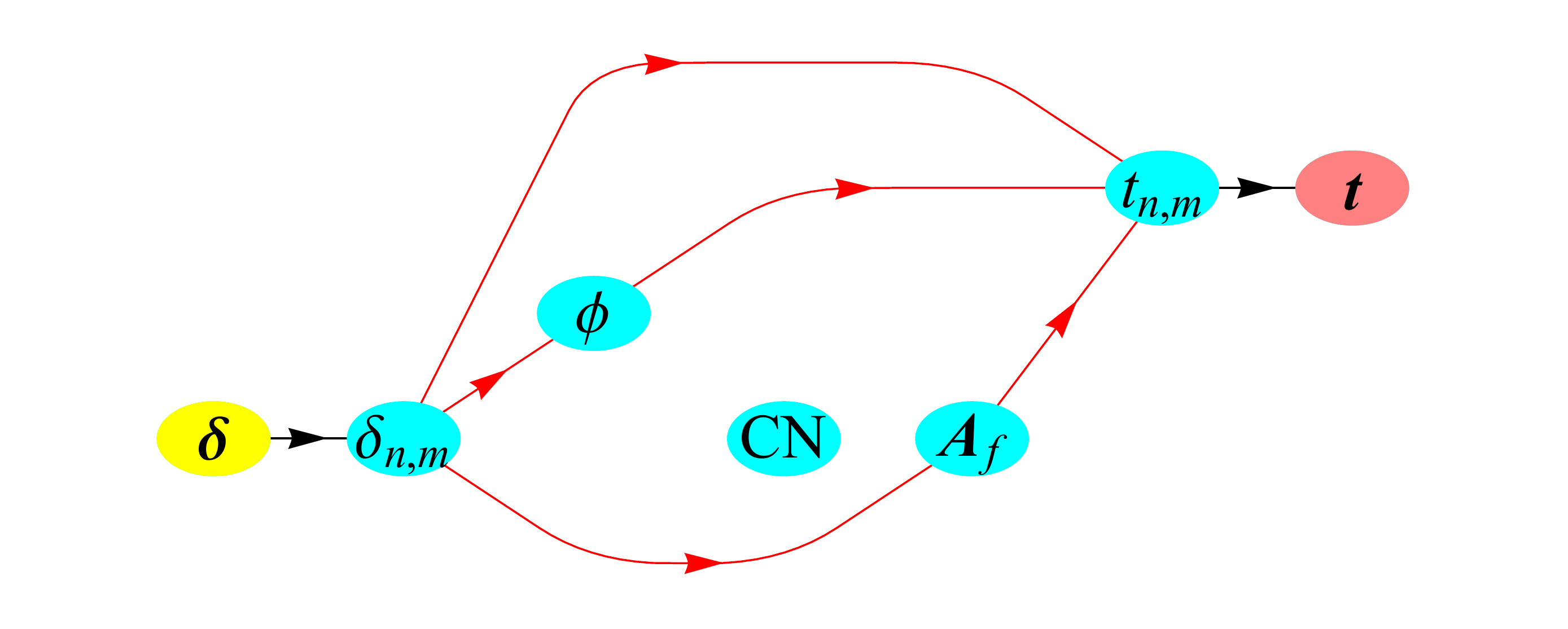}
	}
	\subfigure[The digraph violates Rule No. 4, since there exists a cycle $\phi \rightarrow \tensor{A}_{f} \rightarrow CN \rightarrow \phi$.]{
		\includegraphics[width=0.45\textwidth]{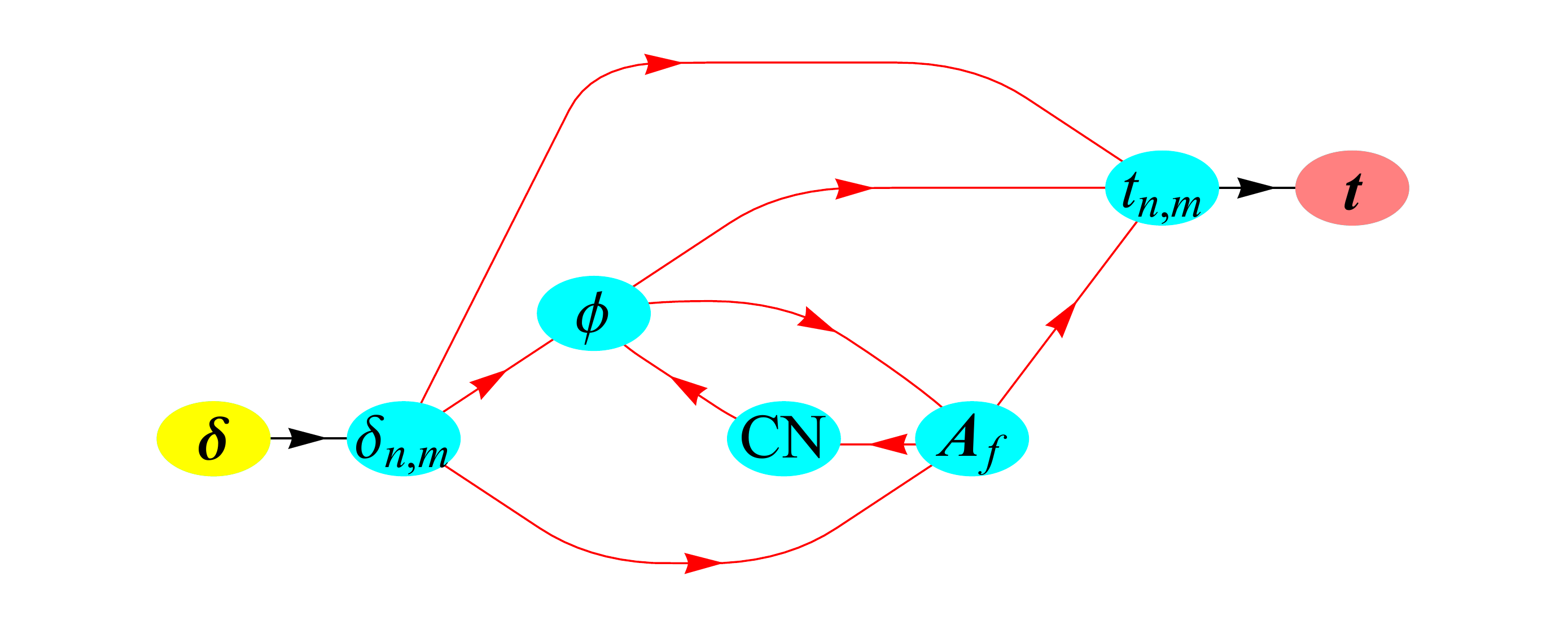}
	}
	\subfigure[The digraph violates Rule No. 5, since $CN$ and $\tensor{A}_{f}$ are not on any paths leading from $\vec{\delta}$ to $\vec{t}$.]{
		\includegraphics[width=0.45\textwidth]{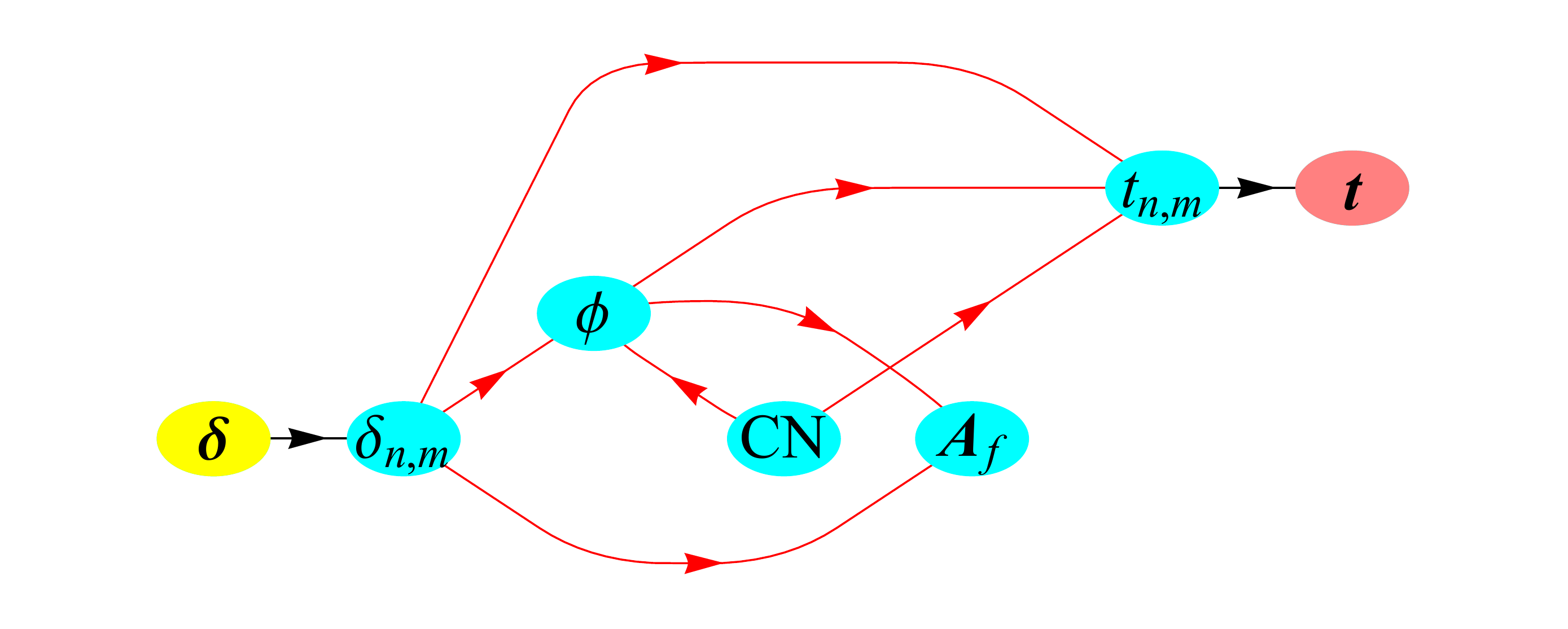}
	}
	\caption{Examples of admissible directed graphs (a-b) and illegal directed graphs (c-d) representing information flow in traction-separation models involving internal physical quantities of porosity $\phi$, coordination number $CN$ and fabric tensor $\tensor{A}_{f}$. The yellow node of separation $\vec{\delta}$ refers to the root node, the pink node of traction $\vec{t}$ refers to the leaf node, and the cyan nodes refer to intermediate nodes. The black arrows refer to "definition" or "universal principles" edges. The red arrows refer to "phenomenological relations" edges.  }
	\label{fig:model_exmple_digraph}
\end{figure}

To sum up, in the digraph representations of traction-separation models, $\vec{\delta}$ is the root and $\vec{t}$ is the leaf, and currently we consider $\vec{q}$ to be a subset of the following set of physical quantities $\{ \delta_{n,m},\ t_{n,m},\ \phi,\ CN,\ \tensor{A}_{f},$ $\ \tensor{A}_{sf},\ d_{a},\ c_{t},\ l_{sp},\ \rho_{g} \}$. 
For the edges, we classify them as either "definitions" (such as $t_{n,m} \rightarrow \vec{t}$, $\vec{\delta} \rightarrow \delta_{n,m}$) which are determined by universal principles in mechanics and should not be modified, or the "phenomenological relations" (such as $\delta_{n,m} \rightarrow \tensor{A}_{f}$, $\phi \rightarrow CN$, $l_{sp} \rightarrow t_{n,m}$) which incorporate material parameters chosen to fit experimental data. 
The latter category of edges provide opportunities for researchers to propose hand-crafted constitutive relations of different degrees of complexities. 
For example, their forms can be linear, quadratic, exponential functions or be approximated by artificial neural networks (ANNs). 
For illustration purposes, we consider a simple digraph of traction-separation models involving only the nodes $\{ \vec{\delta},\ \vec{t},\ \delta_{n,m},\ t_{n,m},\ \phi,\ CN,\ \tensor{A}_{f}\}$. 
Figure \ref{fig:model_exmple_digraph} provides examples of two admissible and two illegal digraph configurations according to the Rules 1-5.

\section{Score system for model evaluation and objective function} \label{sec:objective_score}
% SUN: I suggest that you define the objective function first. Then, in the next section, we simply say that we use the traction-seperation law. 
A score system must be introduced to evaluate the generated directed graphs for constitutive models 
such that the accuracy and credibility in replicating the mechanical behavior of real-world materials can be assessed. 
This score system may also serve as the objective function that defines the rewards for the deep reinforcement learning agent to improve the generated digraphs and resultant constitutive laws. 
In this work, we define 
the score as a positive real-valued function of the range $[0,1]$ which depends on the measures $A_i$ $(i=1,2,3,...,n)$ of $n$ important features of a constitutive model,
\begin{equation}
\text{SCORE} = F(A_1, A_2, A_3, ..., A_n),
\end{equation}
where $0\leq A_i \leq 1$. 
Some features are introduced to measure the performance of a model such as the accuracy and computation speed. 
Other features are introduced to enforce constraints to ensure 
 the admissibility of a constitutive model, such as the frame indifference and the thermodynamics consistency. 
Suppose there are $n_{\text{pfm}}$ measures of performance features $A^{\text{pfm}}_i$ and $n_{\text{crit}}$ measures of critical features $A^{\text{crit}}_i$ in the measure system of constitutive models, the score takes the form,
\begin{equation}
\text{SCORE} = (\prod_{j=1}^{n_{\text{crit}}} A^{\text{crit}}_j) \cdot (\sum_{i=1}^{n_{\text{pfm}}} w_i A^{\text{pfm}}_i),
\end{equation}
where $w_i \in [0,1]$ is the weight associated with the measure $A^{\text{pfm}}_i$, and $\sum_{i=1}^{n_{\text{pfm}}} w_i = 1$. 
In this section, two examples of measures of accuracy $A_{\text{accuracy}}$ and prediction consistency $A_{\text{consistency}}$ are presented. 

\subsection{Accuracy of calibrations and forward predictions}
In this work, the abilities of the models to replicate calibration data and make forward predictions are considered separately. 
Here we introduce a cross-validation procedure in which the dataset used for training the models (e.g. identifying material 
parameters (e.g. \citet{wang2016identifying, liu2016determining}) or adjusting weights of neurons in recurrent neural networks (e.g. \citet{lefik2002artificial, wang2018multiscale}) is mutually exclusive to the testing dataset used to evaluate the quality of blind predictions. 
%We then use the model generated from reinforcement learning to  make predictions and compare the results with the  test data set, which contains data that have not been presented to the models in the calibration phase. 
The details of the generation of these calibration and testing data sets 
using frictional discrete element simulations  are presented in Appendix A. 
Both calibration and blind prediction results are compared against the target data. 
The mean squared error (MSE) commonly used in statistics and also as objective function in machine learning is chosen as the error measure for each data sample $i$ in this study, i.e., 
\begin{equation}
\text{MSE}_{i} = \frac{1}{N_{\text{feature}}} \sum_{j=1}^{N_{\text{feature}}} [\mathcal{S}_{j} (Y_{i_{j}}^{\text{data}}) -\mathcal{S}_{j} (Y_{i_{j}}^{\text{model}})]^2,
\label{eq:mse_data_i}
\end{equation}
where $Y_{i_{j}}^{\text{data}}$ and $Y_{i_{j}}^{\text{model}}$ are the values of the $j$th feature of the $i$th data sample, from target data value and predictions from constitutive models, respectively. 
$N_{\text{feature}}$ is the number of output features. 
$\mathcal{S}_{j}$ is a scaling operator (standardization, min-max scaling, ...) for the output feature $\{Y_{i_{j}}\},\ i \in [1,N_{\text{data}}]$. 

The empirical cumulative distribution functions (eCDFs) are computed for MSE of the entire dataset $\{\text{MSE}_{i}\},\ i \in [1,N_{\text{data}}]$, for MSE of the training dataset $\{\text{MSE}_{i}\},\ i \in [1,N_{\text{traindata}}]$ and for MSE of the test dataset $\{\text{MSE}_{i}\},\ i \in [1,N_{\text{testdata}}]$, with the eCDF defined as \citep{kendall1946advanced},
\begin{equation}
F_{N}(\text{MSE}) = \left \{
\begin{aligned}
&0, &\text{MSE} < \text{MSE}_1,\\
&\frac{r}{N}, &\text{MSE}_{r} \leq \text{MSE} < \text{MSE}_{r+1},\ r = 1,...,N-1,\\
&1, &\text{MSE}_{N} \leq \text{MSE},
\end{aligned}
\right .
\label{eq:ecdf_mse}
\end{equation}
where $N = N_{\text{data}}$, or $N_{\text{traindata}}$, or $N_{\text{testdata}}$, and all $\{\text{MSE}_{i}\}$ are arranged in increasing order. 
A measure of accuracy is proposed based on the above statistics,
\begin{equation}
A_{\text{accuracy}} = \max(\frac{ \log [\max(\varepsilon_{P\%}, \varepsilon_{\text{crit}})] }{\log \varepsilon_{\text{crit}}}, 0),
\label{eq:acc_indicator}
\end{equation}
where $\varepsilon_{P\%}$ is the $P$th percentile (the MSE value corresponding to $P\%$ in the eCDF plot) of the eCDF on the entire, training or test dataset. 
$\varepsilon_{\text{crit}} \ll 1$ is the critical MSE chosen by users such that a model can be considered as "satisfactorily accurate" when $\varepsilon_{P\%} \leq \varepsilon_{\text{crit}}$. 

\subsection{Consistency of accuracy between calibrations and forward predictions}
For the examination of the consistency in model predictions on training data and test data, the K-sample Anderson-Darling (AD) test of goodness-of-fit (gof) is conducted to check whether the eCDFs of training and test data come from the same probability distribution, while this distribution is unspecified \citep{anderson1954test, scholz1987k}. 
It is a non-parametric hypothesis test and determines whether the null hypothesis $H_0$ that the two eCDFs come from the same continuous distribution can be rejected or not, under a chosen significance level $\alpha_{\text{gof}}$. 
The method consists of calculating a normalized AD test statistic, critical values of the AD statistic that depends on the sample sizes, and a $p$-value indicating the approximated significance level at which $H_0$ can be rejected. 
If the $p$-value is smaller than the significance level $\alpha_{\text{gof}}$, the $H_0$ hypothesis is rejected. 
Otherwise there is insufficient evidence to reject $H_0$. 
In this work, we define the following binary measure for the consistency of the MSE distributions, with the significance level $\alpha_{\text{gof}}$, 
\begin{equation}
A_{\text{consistency}} = H^{\alpha_{\text{gof}}} = \left \{
\begin{aligned}
&0,\ & p\text{-value} < \alpha_{\text{gof}},\\
&1,\ & p\text{-value} \geq \alpha_{\text{gof}}.
\end{aligned}
\right .
\label{eq:ADtest_indicator}
\end{equation}
% SUN: You need to talk about why you make the consistant feature A_{consistency} binary 8/23/2018.

\section{Game of the traction-separation law} \label{sec:game}
%SUN: Please make the table ? 6/29
%SUN: I think you need more organization to define the grammar, rule and the operation of the game. For instance, you need to state the full list of ingredients, then add the table. 

Our focus in this paper is primarily on the meta-modeling game invented for generating traction-separation models. Nevertheless, 
similar games can be defined for generating other types of constitutive models based on the ideas presented in this work. 
With the directed graph representations of traction-separation models as presented in Section \ref{sec:graph}, the process of developing a model can be recast as a game of making a sequence of decisions in generating edges between nodes in the digraphs. 
The player of the game can be a human or an AI agent. 
The game starts with an initial "game board" of digraph with predefined nodes and no edge formed between them. 
Each step of the game consists of activating only one edge among all possible choices of edges in the predefined action space, following the predefined rules of the game.   
The game terminates when a complete and admissible digraph following the rules in Section \ref{sec:graph} is established. 
The output models of the game are measured by a score system as presented in Section \ref{sec:objective_score}. 

\begin{figure}[h!]\center
	\subfigure[Initial configuration of the "game board"]{
		\includegraphics[width=0.48\textwidth]{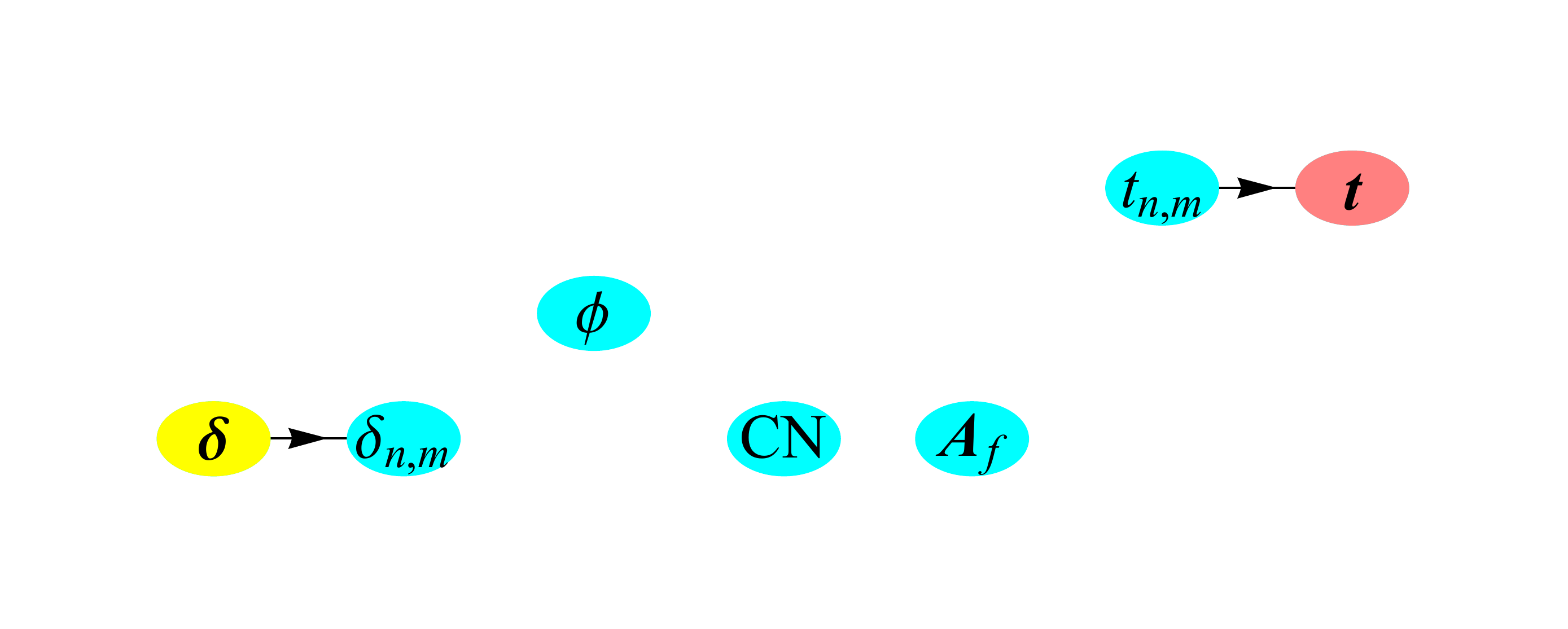}
	}
	\subfigure[All possible actions on the "game board"]{
		\includegraphics[width=0.48\textwidth]{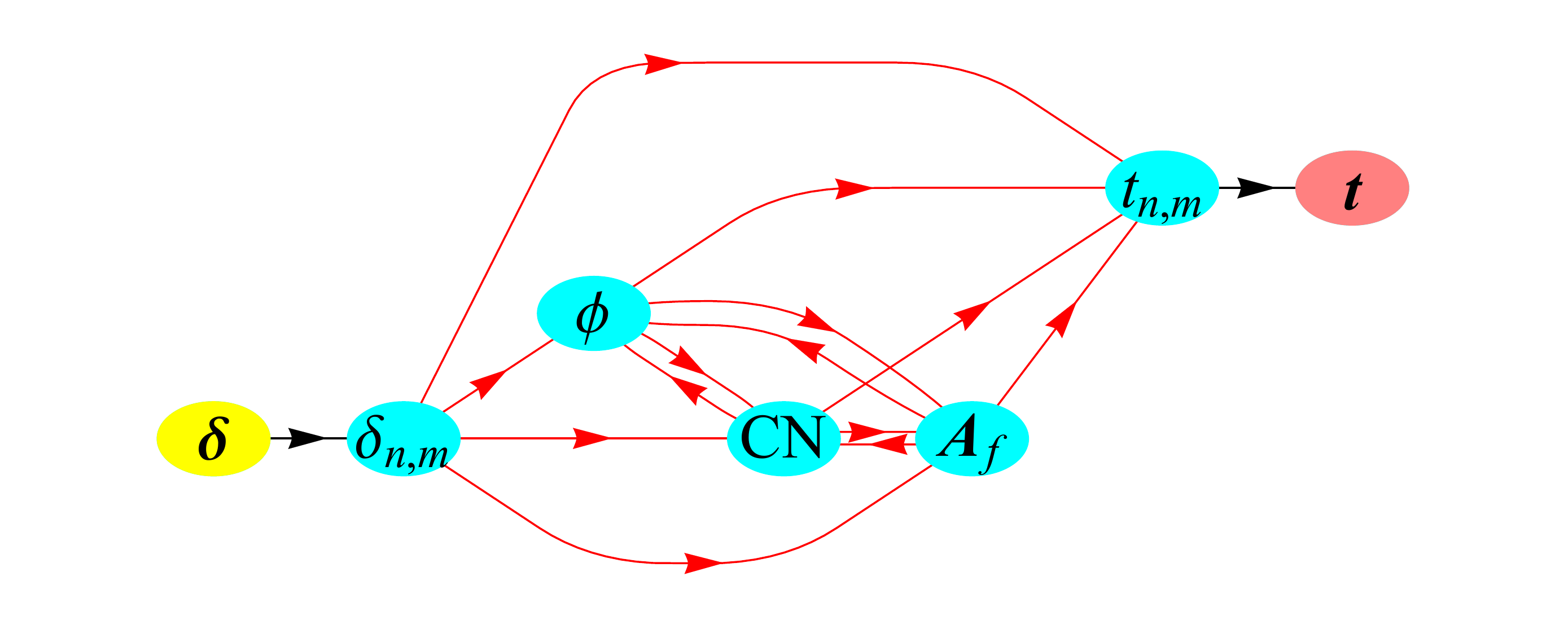}
	}
	\caption{A game of traction-separation model for the digraph example in Figure \ref{fig:model_exmple_digraph}. (a) the initial "board" on which the game is played. (b) All possible actions for picking the edges connecting the nodes are represented by the red arrows.}
	\label{fig:model_game_board}
\end{figure}

The game can be mathematically formalized as a Markov decision process. 
The human or AI agent observes the state of the game $s_t$ at the current step $t$ from the game environment (the directed graph that represents a constitutive model) in the form of a vector of binaries indicating the on/off status of each valid edge choices in the action space. 
The agent takes an action $a_t$ on the game environment in the form of an integer indicating the next edge to switch on in the action space. 
The action $a_t$ is sampled from a vector of probabilities $\pi(s_t)$ of taking each valid action from the state $s_t$. 
Consequently, the state of the game becomes $s_{t+1}$ at the next step $t+1$. 
The agent also receives a reward $r_{t+1}$ for the action $a_t$ of taking the game state from $s_t$ to $s_{t+1}$. 
Each policy applied in a complete gameplay produces a particular trajectory $s_0$, $a_0$, $r_1$, $s_1$, $a_1$, $r_2$, ..., $a_{t-1}$, $r_t$, $s_{t}$, $a_{t}$, ..., $a_{T-1}$, $r_T$, $s_T$. 
Once a complete constitutive model is generated, the model score is evaluated. 
The final reward $r_T$ is defined as: if the current score is higher than the average score of models from a group of already played games by the agent, then the current game wins and $r_T=1$, otherwise, the current game loses and $r_T=-1$. 
The average score can be initialized to 0 for the first game. 

Note that the reward is only available at the end of the game, similar to the game of Chess and Go.
$r_T$ is known according to the score of the generated model. 
The previous rewards $r_{t<T}$, however, can only be estimated. 
For a human agent, both rewards $r_{t<T}$ and move probabilities $\pi(s)$ come from "intuition" gained during many constitutive modeling practices. 
An experienced human modeler estimates the rewards and probabilities more accurately and hence more likely generates better constitutive models.
For an AI agent, $r_{t<T}$ and $\pi(s)$ are approximated by hand-crafted mathematical functions or recently neural networks as in deep reinforcement Q-learning. 
They are estimated based on the expected game reward of taking action $a$ from state $s$ (Q-value) $Q(s,a)$ and the value of current state $v(s)$. 
The above-mentioned important quantities for mathematical descriptions of the gameplays are summarized in Table \ref{tab:constitutive_game_define}. 
Moreover, the constitutive modeling game is compared side-by-side with the game of Chess more familiar to the public in Table \ref{tab:chess_constitutive_game_compare}, in the aspects of the board to play on, the permitted actions to execute, the criteria for wining the game, etc. 
\begin{table}[t]
	\centering
	\begin{tabular}{| l | l |}
		\hline
		Environment & Benchmark training and test data, idealized multigraph for constitutive models\\ \hline
		Agent & Human or AI \\ \hline
		State $s$ & A list of binaries indicating the on/off status of each valid edge choice \\ \hline
		Action $a$ & An integer indicating the next edge to switch on from the current game state \\ \hline
		Reward $r$ & Win (1) / loss (-1) according to the score of the constitutive model in Section \ref{sec:objective_score} \\ \hline
		$\pi(s,a)$ & Probability of taking action $a$ at state $s$ \\ \hline
		$v(s)$ & Expected reward of state $s$ \\ \hline
		Q-value $Q(s,a)$ & Expected reward from taking action $a$ at state $s$ \\ \hline
	\end{tabular}
	\caption{Key ingredients of the game of constitutive models in directed graph.}
	\label{tab:constitutive_game_define}
\end{table}

\begin{table}[t]
	\centering
	\begin{tabular}{| c | p{5cm} | p{6cm} |}
		\hline
		& Game of Chess & Game of constitutive modeling in directed graph\\ \hline
		Definition of game & Make a sequence of decisions to maximize the probability to win & Make a sequence of decisions to maximize the score of the constitutive model\\ \hline
		Game board & 8$\times$8 grid & Directed graph with predefined nodes of physical quantities and edges of definition or universal principles\\ \hline
		Game state & Configuration of chess pieces on the board & Configuration of directed graph representing the constitutive model\\ \hline
		Game action & Move chess pieces & Select among modeling choices. For instance
		\begin{enumerate}
			\item  What physical quantities are included?
			\item  How physical quantities are linked?
			\item What are the edges between physical quantities?
		\end{enumerate}
		\\ \hline
		Game rule & Restrictions on chess piece movements & Universal principles 
		
		Rules in Section \ref{sec:graph} 
		
		Specific restrictions on edge choices\\ \hline
		Game reward & Win, draw or loss (discontinuous) & Win or loss (discontinuous) from comparison of model scores (continuous)\\ \hline
		Reward evaluation & Only available at the end & Only available at the end\\ \hline
	\end{tabular}
	\caption{Comparison of the essential definitions between the game of Chess and the game of constitutive modeling in directed graph.}
	\label{tab:chess_constitutive_game_compare}
\end{table}

\begin{figure}[h!]\center
	\includegraphics[width=1\textwidth]{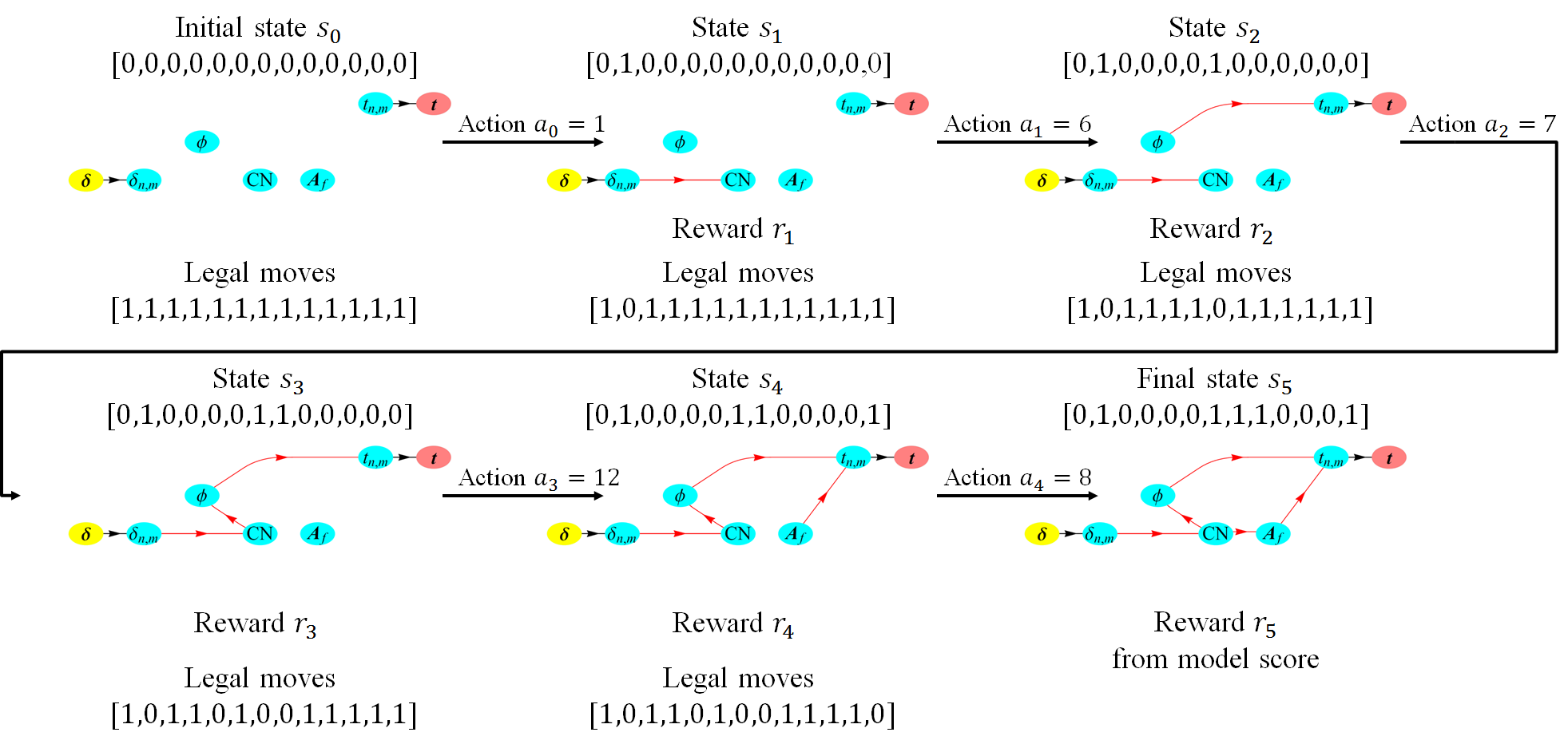}
	\caption{A gameplay example formalized as a Markov decision process ($s_0$, $a_0$, $r_1$, $s_1$, $a_1$, $r_2$, $s_2$, $a_2$, $r_3$, $s_3$, $a_3$, $r_4$, $s_4$, $a_4$, $r_5$, $s_5$) for the digraph game in Figure \ref{fig:model_game_board}. The states are lists of binaries for 13 edges $[ \delta_{n,m} \rightarrow \phi,\ \delta_{n,m} \rightarrow CN,\ \delta_{n,m} \rightarrow \tensor{A}_f,\ \delta_{n,m} \rightarrow t_{n,m},\ \phi \rightarrow CN,\ \phi \rightarrow \tensor{A}_f,\ \phi \rightarrow t_{n,m},\ CN \rightarrow \phi,\ CN \rightarrow \tensor{A}_f,\ CN \rightarrow t_{n,m},\ \tensor{A}_f \rightarrow \phi,\ \tensor{A}_f \rightarrow CN,\ \tensor{A}_f \rightarrow t_{n,m}]$. The actions $a$ are integers $\in [0,12]$ (the list indices start from 0) indicating the next edge ID to activate. The legal moves are lists of binaries indicating whether the edges are allowed to be activated next. Final reward $r_5$ is determined by the model score evaluated at the end of the game. $r_{1-4}$ are only estimated by "intuitions" on whether the current policy can lead to a win or not, until $r_5$ is known. Note that the Markov decision process leading to the final digraph configuration $s_5$ is not unique.}
	\label{fig:model_gamestep_example}
\end{figure}

For illustration purposes, we provide a simple game example for the digraph presented in Figure \ref{fig:model_exmple_digraph}, which only involves the nodes $\{ \vec{\delta},\ \vec{t},\ \delta_{n,m},\ \ t_{n,m},\ \phi,\ CN,\ \tensor{A}_{f}\}$. 
Figure \ref{fig:model_game_board} presents the "initial game board" and all possible edges choices in the current game definition. 
The configuration of the digraph, or the state of the game, can be totally described by a list of binaries for 13 edges $[ \delta_{n,m} \rightarrow \phi,\ \delta_{n,m} \rightarrow CN,\ \delta_{n,m} \rightarrow \tensor{A}_f,\ \delta_{n,m} \rightarrow t_{n,m},\ \phi \rightarrow CN,\ \phi \rightarrow \tensor{A}_f,\ \phi \rightarrow t_{n,m},\ CN \rightarrow \phi,\ CN \rightarrow \tensor{A}_f,\ CN \rightarrow t_{n,m},\ \tensor{A}_f \rightarrow \phi,\ \tensor{A}_f \rightarrow CN,\ \tensor{A}_f \rightarrow t_{n,m}]$ (The edges $\vec{\delta} \rightarrow \delta_{n,m}$ and $t_{n,m} \rightarrow \vec{t}$ are definitions and always active). 
The list also represents the entire action space.
The action $a$ is an integer $\in [0,12]$ indicating the next edge ID to activate in the list. 
The legal moves at the current game state are represented by a list of 13 binaries indicating whether the corresponding edges are allowed to be activated for the next action step. 
The rule of the legal moves are as follows: (1) if one edge has already been selected, it is excluded from the selection of actions; (2) if an edge between two intermediate nodes has been selected, the other edge involving these two nodes but with opposite direction is also excluded (e.g., The edges $\phi \rightarrow CN$ and $CN \rightarrow \phi$ are mutually exclusive); (3) the resultant digraph must obey the rules in Section \ref{sec:graph}. 
Figure \ref{fig:model_gamestep_example} provides a gameplay example of the constitutive modeling game in Figure \ref{fig:model_game_board}, with mathematical representations of game states, actions and legal actions, as well as the Markov decision process.

The score evaluation (Section \ref{sec:objective_score}) requires model calibration on training data, and forward predictions on test data. 
The procedure for score evaluation is as follows. 
Once the final digraph configuration is determined, all paths (information flows) leading from $\vec{\delta}$ to $\vec{t}$ and all predecessors for each node in the paths are identified using the graph theory (software package NetworkX). 
Secondly, the predecessor nodes for the terminal node $\vec{t}$ within these paths are identified. 
Recursively going upstream along all the information flows, the predecessors for these nodes are identified, until the final predecessor node is the start node $\vec{\delta}$ only. 
All the predecessor-successor node pairs can be connected by either mathematical equations frequently used in handcrafted constitutive models (linear, quadratic, exponential, power law, etc.) or artificial neural networks. 
In this work, we take the advantage of the flexibility of ANNs that they are universal function approximators to continuous functions of various complexity on compact subsets of $R^n$ (Universal approximation theorem, \citep{hornik1989multilayer}). 
Moreover, a special type of ANN, recurrent neural network (e.g., long short-term memory LSTM \citep{hochreiter1997long}, gated recurrent units GRU \citep{cho2014learning, chollet2015keras}), can capture the function of a time series of inputs, which is ideal for replicating the path-dependent material behavior. 
Hence, we only focus on ANN edges, without loss of generality of the meta-modeling games. 
The hybridized constitutive models with both mathematical equation edges and ANN edges will be studied in a separate research. 
The predecessor-successor node pairs are also inputs and outputs of all ANNs involved in the constitutive model. 
For example, there are two paths in the final digraph $s_5$ in Figure \ref{fig:model_gamestep_example}: $\{\vec{\delta} \rightarrow \delta_{n,m} \rightarrow CN \rightarrow \tensor{A}_f \rightarrow t_{n,m} \rightarrow \vec{t}\}$ and $\{\vec{\delta} \rightarrow \delta_{n,m} \rightarrow CN \rightarrow \phi \rightarrow t_{n,m} \rightarrow \vec{t}\}$. 
Then the three required ANNs are, represented as input-output pairs, $[\delta_{n,m} \rightarrow CN]$, $[CN \rightarrow \phi, \tensor{A}_f]$ and $[\phi, \tensor{A}_f \rightarrow t_{n,m}]$. 
The parameters in each ANN are calibrated with training data of the input and output features using back propagations. 
The final output of $\vec{t}$ is predicted by executing consecutively the ANNs following the established paths from $\vec{\delta}$ to $\vec{t}$ in the directed graph. 
In the numerical examples of this paper, the same neural network architecture is used for all ANNs: two hidden layers of 32 GRU neurons in each layer, and the output layer is a dense layer with linear activation function. 
All input and output data are preprocessed by standard scaling using mean values and standard deviations \citep{scikit-learn}. 
Each input feature considers its current value and 19 history values prior to the current loading step. 
Each ANN is trained for 1000 epochs using the Adam optimization algorithm \citep{kingma2014adam}, with batch size of 256. 

\section{Deep reinforcement learning for generating constitutive laws} \label{sec:gameplay}
\begin{figure}[h!]\center
	\includegraphics[width=1.0\textwidth]{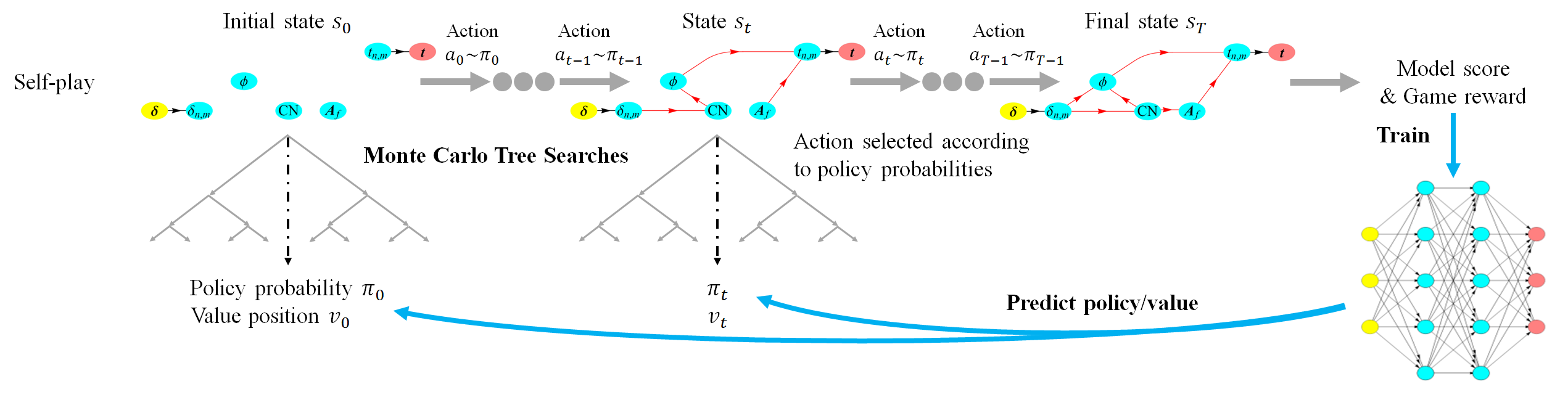}
	\caption{Self-play reinforcement learning of traction-separation law.}
	\label{fig:selfplay_learn}
\end{figure}

%SUN: Take a look at the Stanford slides. 6/29
With the game of constitutive modeling completely defined, a deep reinforcement learning (DRL) algorithm is employed as a guidance of taking actions in the game to maximize the final model score (Figure \ref{fig:selfplay_learn}). 
This tactic is considered one of the key ideas leading to the major breakthrough in AI playing 
 the game of Go (AlphaGo Zero) \citep{silver2017mastering}, Chess and shogi (Alpha Zero) \citep{silver2017masteringchess} and many other games. 
The learning is completely free of human interventions. 
It does not need previous human knowledge in traction-separation model as a starter database. 
The AI agent simply learns to improve from a number of games it played and from the corresponding model scores and game rewards, even if the initially generated digraph configurations make very little sense for a traction-separation model. 
Moreover, during the self-plays and training, no human guidance is needed. 

A (deep) neural network $f_{\theta}$ with parameters $\theta$ (weights, bias, ... of the artificial neurons) takes in the current configuration of the directed graph of the constitutive law $s$ and outputs a policy vector $\vec{p}$ with each component $p_a = p(s,a)$ representing the probability of taking the action $a$ from state $s$, as well as a scaler $v$ estimating the expected score of the constitutive law game from state $s$, i.e.,
\begin{equation}
(\vec{p}, v) = f_{\theta}(s).
\end{equation}
These outputs from the policy/value network guide the game play from the AI agent. 

\begin{figure}[h!]\center
	\includegraphics[height=6cm, width=\textwidth]{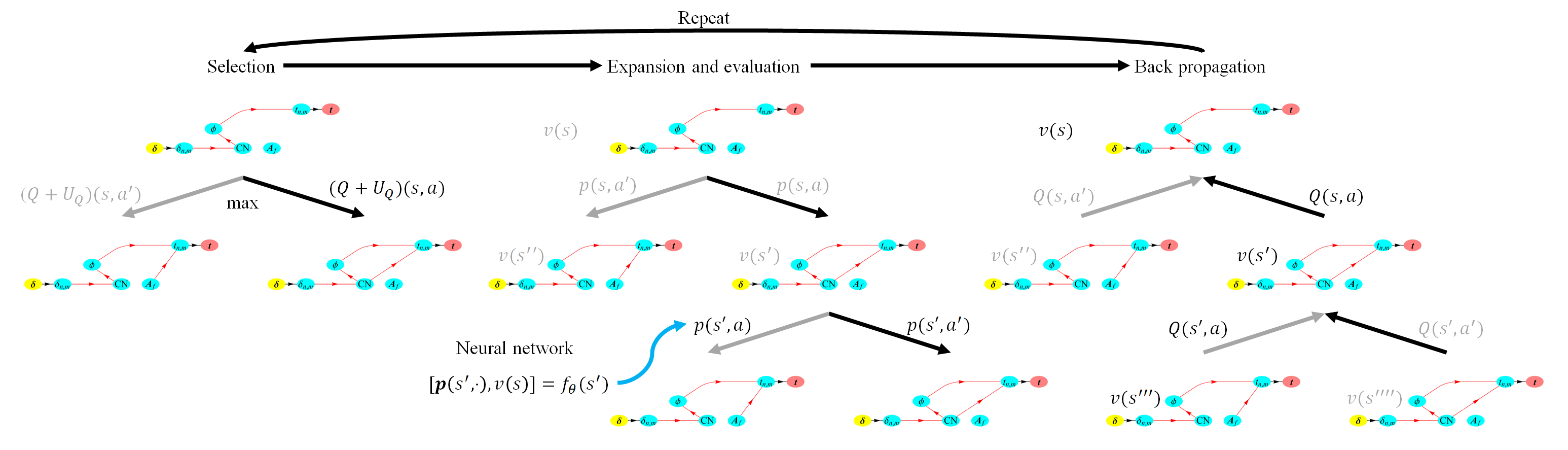}
	\caption{Monte Carlo Tree Search (MCTS) in a game of constitutive models (figure design adopted from \citep{silver2017mastering}). A sequence of actions are selected from the root state $s^0$, each maximizing the upper confidence bound $Q(s,a) + U_Q(s,a)$. The leaf node $s^l$ is expanded and its policy probabilities and position value are evaluated from the neural network $(\vec{p}(s^l), v(s^l))=f_{\theta}(s^l)$. The action values $Q$ in the tree are updated from the evaluation of the leaf node. Finally the search probability $\vec{\pi}(s^0)$ for the root state $s^0$ is returned to guide the next action in self-play.}
	\label{fig:mcts_graph}
\end{figure}

At each state $s$, the action to take is sampled from an action probability $\vec{\pi}(s)$. 
This probability is based on the policy $\vec{p}$ predicted from the neural network enhanced by a Monte Carlo Tree Search (MCTS) (\citep{browne2012survey}). 
The search tree is composed of nodes representing states $s$ of the game, and edges representing permitted actions $a$ from $s$. 
Each edge $(s,a)$ possesses a list of statistics $[N(s,a), W(s,a), Q(s,a)]$, where $N(s,a)$ is the number of visits to the edge during MCTS search, $W(s,a)$ is the total action value and $Q(s,a) = \frac{W(s,a)}{N(s,a)}$ is the mean action value. 
The search procedure consists of firstly a recursive selection of a sequence of optimal actions $a^0,a^1,a^2,...$ leading to the corresponding child states $s^1,s^2,s^3,...$, starting from the root state $s^0$, until a leaf node of state $s^l$ (that has never been encountered before in the search) is reached. 
The criteria for selection from a state $s$ is that the action $a$ maximizes the upper confidence bound $U(s,a)$ of the Q-value, among all valid actions. 
The upper bound is defined as
\begin{equation}
U(s,a) = Q(s,a) + U_Q(s,a) = Q(s,a) + c_{puct} p(s,a) \frac{\sqrt{\sum_b N(s,b)}}{1+N(s,a)}.
\end{equation}
where $c_{puct}$ is a parameter controlling the level of exploration. 
If $s^l$ is not a terminal state that ends the game, then its $\vec{p}(s^l)$ and $v(s^l)$ are predicted from the policy/value neural network $f_{\theta}(s^l)$. 
The search tree is expanded and the statistics for each edge $(s^l, a)$ is initialized to $[N(s,a)=0, W(s,a)=0, Q(s,a)=0]$. 
Otherwise, $v(s^l)$ is equal to the final reward of the constitutive modeling game. 
Finally, $v(s^l)$ is propagated back to the parent states $\{s^0,s^1,s^2,...s^l\}$ and actions $\{a^0,a^1,a^2,...a^{(l-1)}\}$ traversed during the seach. 
Their statistics are updated as 
\begin{equation}
N(s^i,a^i) = N(s^i,a^i) + 1,\ W(s^i,a^i) = W(s^i,a^i) + v(s^l),\ Q(s^i,a^i) = \frac{W(s^i,a^i)}{N(s^i,a^i)},\ \text{for all}\ i < l.
\end{equation} 
The MCTS procedure is repeated a number of times. 
The searches in MCTS eventually yield a vector of search probabilities $\vec{\pi}(s^0)$ recommending actions to take from the root position $s^0$. 
$\vec{\pi}(s^0)$ is proportional to the exponentiated visit count for each edge, i.e.,
\begin{equation}
\vec{\pi}(s^0, a) = \frac{N(s^0,a)^{-\tau}}{\sum_bN(s^0,b)^{-\tau}},
\end{equation}
where $\tau$ is a positive temperature parameter that also controls the level of exploration. 
The MCTS algorithm for the game of constitutive models is illustrated in Figure \ref{fig:mcts_graph}. 

During one episode of self-play by the AI agent, the above MCTS algorithm is executed for each state $s_t$ in the sequence of encountered states $\{s_0, s_1, s_2,..., s_{T-1}\}$. 
The root node $s^0$ of the search tree is set to $s_t$ as the game progresses to the state $s_t$. 
All child nodes and their statistics constructed in the MCTS for the prior game states are preserved. 
The training data for the neural network consists of $(s_t, \vec{\pi}_t, z_t)$ obtained from a number of full plays of the constitutive law game guided by the aforementioned reinforcement learning algorithm. 
$\vec{\pi}_t$ is the estimation of policy after performing MCTS from state $s_t$ and $z_t$ is the reward of the generated constitutive model at the end of the game $s_T$. 
The loss function to be minimized by adjusting parameters $\theta$ using back propagation is,
\begin{equation}
\text{l} = \sum_t (v(s_t) - z_t)^2 + \sum_t \vec{\pi}_t \log[\vec{p}(s_t)],
\end{equation} 
which is the combination of mean squared errors in game reward (1 or -1) and cross-entropy losses in policy probabilities. 
Hence, accordingly, the activation functions for the output layer are the hyperbolic tangent function $\text{tanh}(x) = \frac{e^{2x}-1}{e^{2x}+1}$ and the softmax function \citep{nasrabadi2007pattern}. 
The procedure of DRL guided self-plays and the following training of the network $f_{\theta}$ is iterated until the score of the generated directed graph of the constitutive model does not improve. 

%$Q(s,a)$ is the estimation of the reward of taking action $a$ at state $s$. 
%The loss function for Q-value learning is derived from the Bellman equation
%\begin{equation}
%\text{loss} = (r+\gamma \max_{a'} Q(s',a') - Q(s,a))^2.
%\end{equation}

\section{Numerical Experiments and Applications} \label{sec:numexperiment}
In this section, we present two traction-separation modeling games with different digraph complexities to demonstrate the intelligence, robustness and efficiency of the deep reinforcement learning algorithm on improving the accuracy and consistency of the generated traction-separation models through self-plays. 
In both examples, sub-scale discrete element simulations (DEM) are used to generate synthetic benchmark data for model calibrations and blind prediction evaluations. 
The procedure of database generation from a pre-consolidated representative volume element (RVE) of a frictional contact material is described in Appendix A. 
The third numerical example presents a multiscale finite element simulation where the previously auto-generated optimal model is used as a scale-bridging constitutive model for pre-existing interface. 

\subsection{Numerical Experiment 1: Determining optimal physical relationships for traction-separation laws}  
In the first example, our goal is to test the DRL algorithm and see whether it can determine the 
optimal topological relations among microstructural physical quantities of porosity $\phi$, coordination number $CN$ and fabric tensor $\tensor{A}_{f}$. 
In \citet{wang2018multiscale}, the authors use domain expertise, i.e., knowledge from previous literature on fabric tensor and critical state theory
to deduce that the porosity and fabric tensor can be used as state variables to improve the forward prediction accuracy of 
the traction-separation law (cf. \citet{fu2011fabric,  li2011anisotropic, sun2013unified, wang2016semi}).  
In this work, we do not make any assumption or introduce any interpretation to the meta-modeling computer agent. 
Instead, we simply make a number of physical quantities measured from discrete element simulations available as vertices in the directed graph but do not 
introduce any relation (edge) manually. 
In other words, the edge set that represents the relations of the physical quantities is self-discovered by the computer agent  from the reinforcement learning without any human intervention. 
We document our training procedure and analyze the performance of the models generated by the meta-modeling approach. 

The directed graphs, states, actions, rewards and game rules of the modeling game have been defined in Sections \ref{sec:graph} and \ref{sec:game}, and illustrated in Figures \ref{fig:model_exmple_digraph}, \ref{fig:model_game_board} and \ref{fig:model_gamestep_example}. 
The action space is of dimension 13. 
Through exhaustive plays of the game, the authors count 3200 possible game states, among which 591 states represent complete and admissible directed graph configurations according to the game rules. 
The model score is defined as:
\begin{equation}
\text{SCORE} = 0.45*A^{\text{calibration}}_{\text{accuracy}} + 0.45*A^{\text{prediction}}_{\text{accuracy}} + 0.1*A_{\text{consistency}},
\label{eq:game1_score}
\end{equation}
where $P\%=90\%$ and $\varepsilon_{\text{crit}}=1e^{-6}$ for accuracy evaluations and $\alpha_{\text{gof}}=1\%$ for consistency evaluations. 
The training data for model calibration contains 50 loading cases, and the test data for forward prediction evaluation contains 150 loading cases. 

\begin{figure}[h!]\center
	\subfigure[Violin plots of the density distribution of model scores in each DRL iteration]{
		\includegraphics[width=0.45\textwidth]{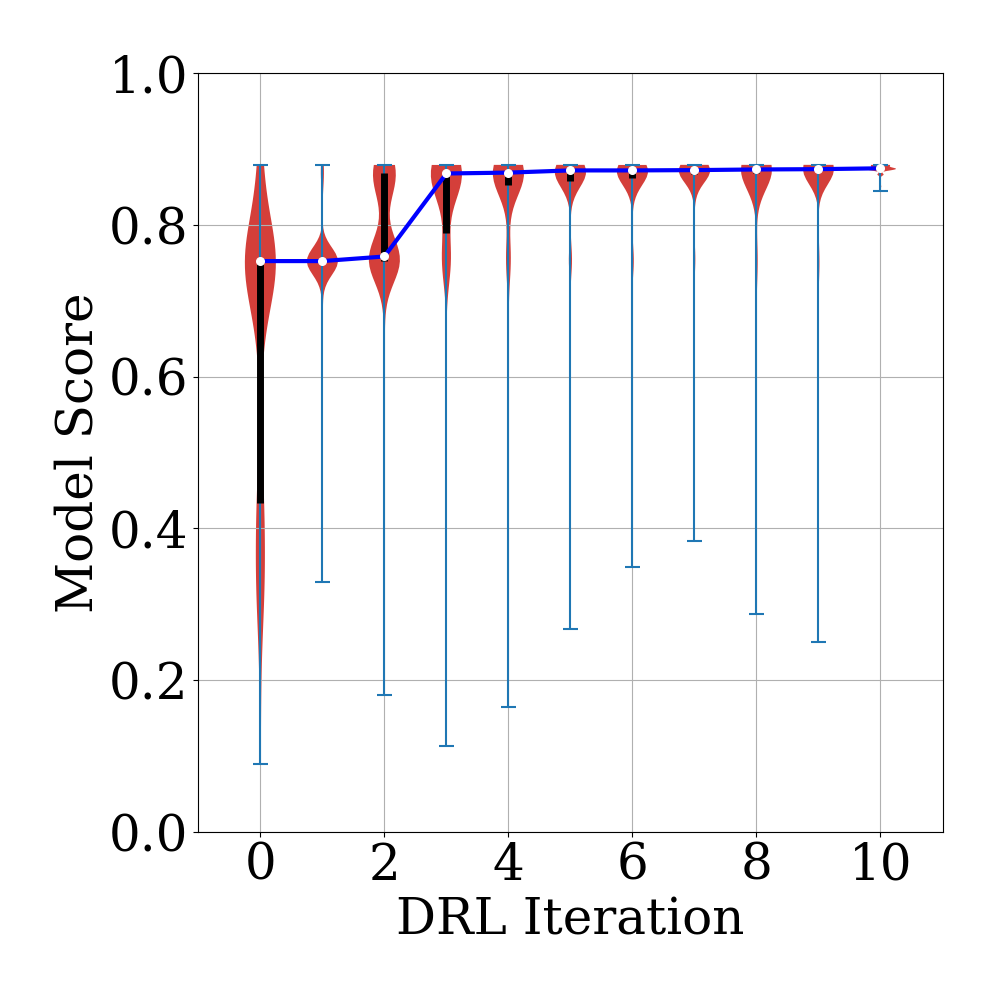}
	}
	\subfigure[Mean value and $\pm$ standard deviation of model score in each DRL iteration]{
		\includegraphics[width=0.45\textwidth]{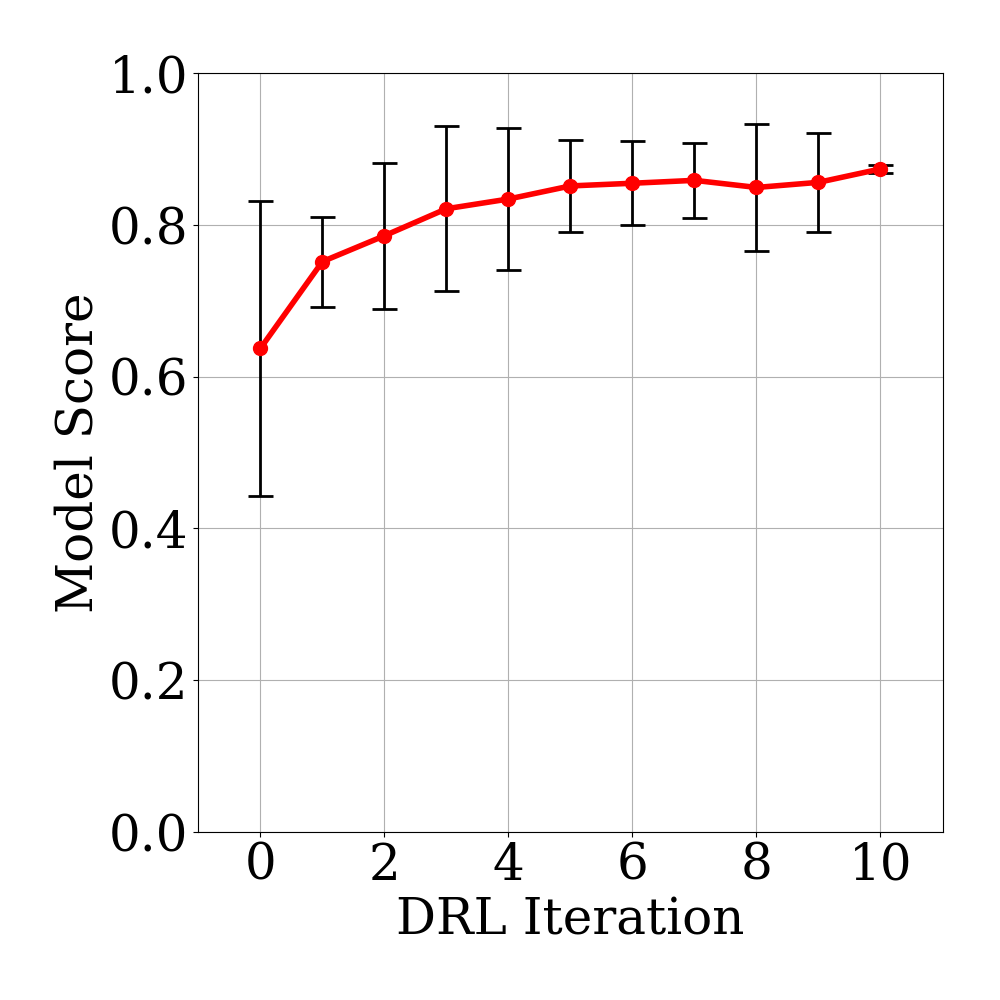}
	}
	\caption{Statistics of the model scores in deep reinforcement learning iterations from 20 separate runs of the DRL procedure for Numerical Experiment 1. Each DRL procedure contains ten iterations 0-9 of "exploration and exploitation" (by setting the temperature parameter $\tau=1.0$) and a final iteration 10 of "competitive gameplay" ($\tau=0.01$). Each iteration consists of 20 games. (a) Violin Plot of model scores played in each DRL iteration. The shade area represents the density distribution of scores. The white point represents the median. The thick black bar represents the interquartile range between 25\% quantile and 75\% quantile. The maximum and minimum scores played in each iteration are marked by horizontal lines. (b) Mean model score in each iteration and the error bars mark $\pm$ standard deviation.}
	\label{fig:game1_statistics}
\end{figure}

The DRL meta-modeling procedure contains 10 iterations of "exploration and exploitation" of game strategies, by setting the temperature parameter $\tau$ to 1. 
Then an iteration of "competitive gameplay" ($\tau=0.01$) is conducted to showcase the performance of the final trained AI agent. 
Each iteration consists of 20 self-plays of the game. 
Each game starts with a randomly initialized neural network for the policy/value predictions, and each play step require 20 MCTS simulations. 
Then the play steps and corresponding final game rewards are append to the set of training examples for the training of the policy/value network. 
Due to the randomness of the initialized neural network and the MCTS search for each play step, each run of the DRL algorithm may yield different game play results, concerning the starting game policy, the speed of improvement, and the converged optimal policy. 
In this numerical example, the DRL procedure is repeated 20 times and all model scores that the AI agent played during the iterations are recorded. 
Hence for each iteration, there are 400 gamplays for statistical analysis (20 gamplays X 20 repeated procedures). 
The gameplay results are presented in Figure \ref{fig:game1_statistics}. 

\begin{figure}[h!]\center
\includegraphics[width=1.0\textwidth]{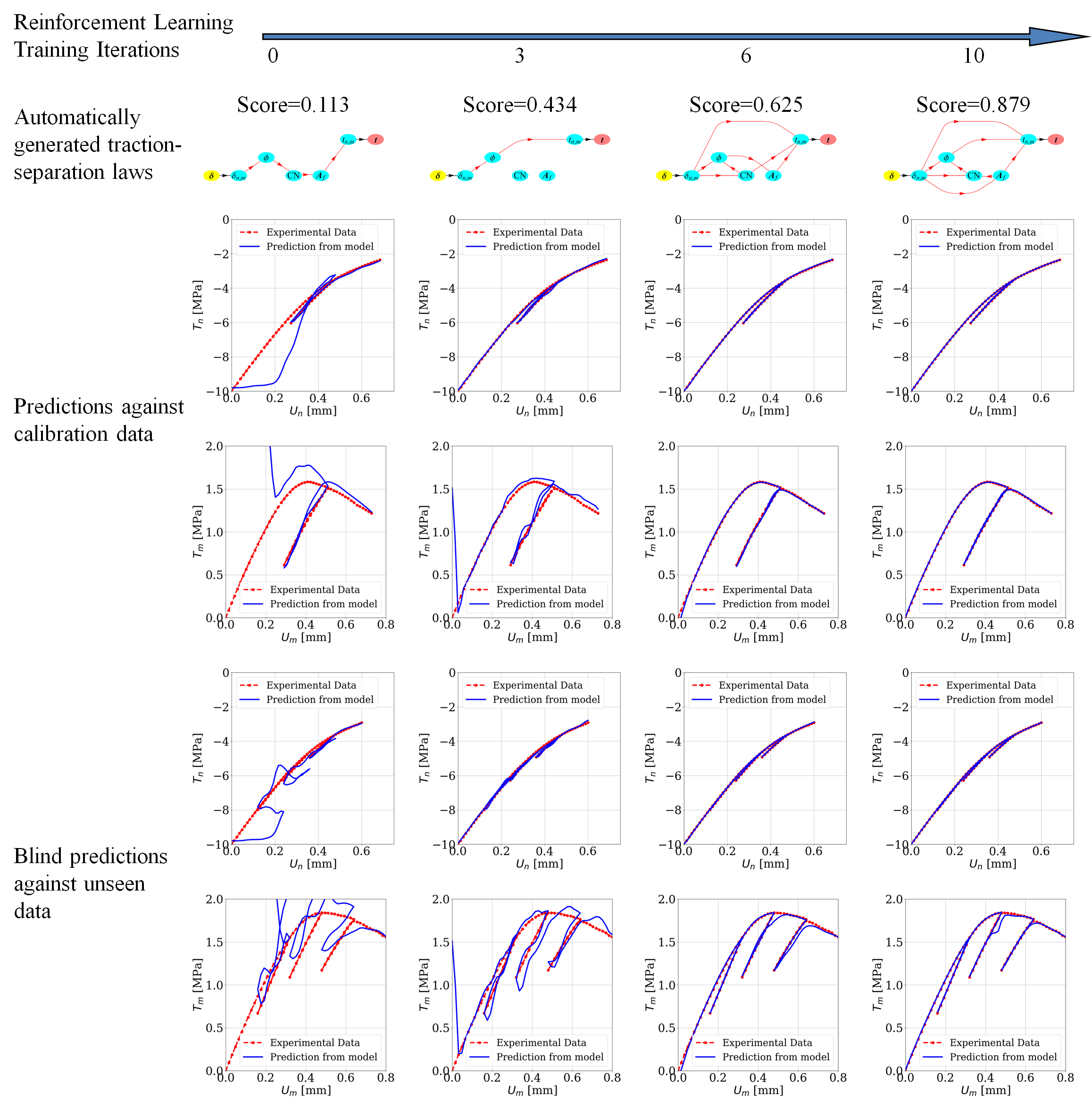}
\caption{Knowledge of directed graphs of traction-separation models learned by deep reinforcement learning in Numerical Experiment 1. Four representative digraph games played during the DRL iterations and their prediction accuracy against training and test data are presented.}
\label{fig:model_drl_learn}
\end{figure}

\begin{figure}[h!]\center
	\includegraphics[width=1.0\textwidth]{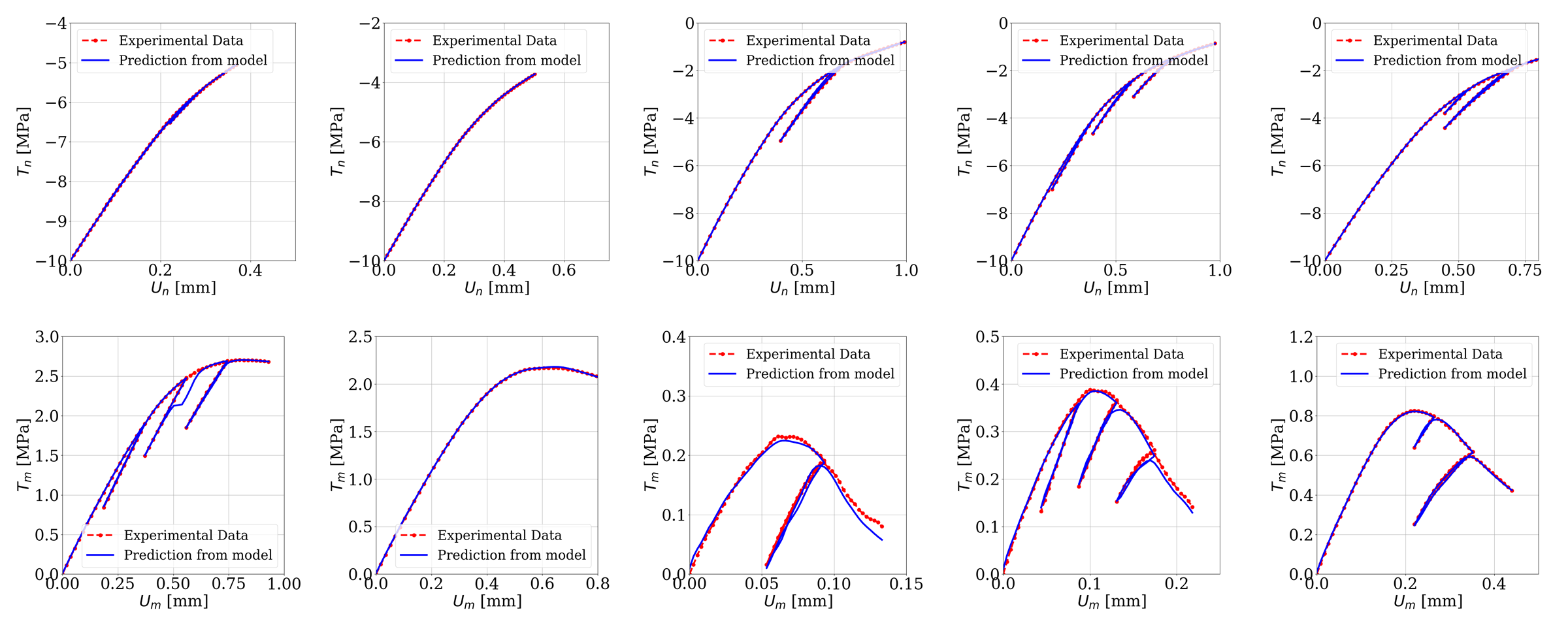}
	\caption{Five examples of blind predictions from the optimal digraph configuration (the 4th digraph in Figure \ref{fig:model_drl_learn}) against unseen data among test database of 150 loading cases.}
	\label{fig:model_drl_predicts}
\end{figure}

At the first DRL iteration, the AI agent only knows the rule of the game without human knowledge on neither which physical quantities are essential in predicting the tractions nor how they should be connected. 
The AI just plays with trial-and-error following strategies guided by random initial neural network and MCTS. 
This lack of gameplay knowledge can be seen from the widely spread density distribution of model scores between maximum and minimum scores, large interquartile range between 25\% and 75\%, and the large standard deviation (Figure \ref{fig:game1_statistics}). 
In the subsequent iterations, the AI plays with increasing knowledge of game play reinforced by the ultimate game rewards, and it shows intelligence in keep playing games with better outcomes. 
This is shown by the increase in median and average of scores, the narrowing of interquartile range and the migration of the density distribution towards higher scores. 
The automatic learning is very efficient. 
Statistically, after 5 iterations (100 games out of the total 591 possible game outcomes), the scores already concentrate around the maximum. 
Few bad games could be played, since the AI is still allowed to explore different game possibilities to avoid convergence to local maximum. 
The strength of the AI agent after 10 iterations is tested by suppressing the "exploration plays", and the outcome game scores show outstanding performances. 
Figure \ref{fig:model_drl_learn} illustrates the improvement of knowledge of traction-separation constitutive modeling by four representative digraph games played during the DRL iterations. 
The traction predictions from the resultant constitutive models are compared against both training data and unseen test data. 
In addition, five examples of blind predictions from the optimal digraph configuration (the 4th digraph in Figure \ref{fig:model_drl_learn}) obtained in this game are shown in Figure \ref{fig:model_drl_predicts}. 

\subsection{Numerical Experiment 2: Data-driven discovery for enhancement of traction-separation laws} 
In the second example, we consider another common scenario in which we attempt to convert qualitative observations into quantitative predictions with the help of the reinforcement learning algorithm as a tool for augmented intelligence. 
The need to interpret observations of mechanisms into predictions is one of the oldest problems in 
constitutive modeling \citep{pastor2011computational}. 
For instance, the observation that yielding depends on the amount of normal traction leads to the 
Mohr-Coulomb yield criterion \citep{timoshenko1953history}. The evolution of fabric tensor has been incorporated into the hardening law
and the plastic flow rule to capture the induced anisotropy and critical state of sand \citep{mehrabadi1982statistical,  dafalias2004simple, dafalias2004sand}. 
However, recent advancements on the application of graph theory as well as the experimental techniques such as micro-CT imaging have 
revealed many geometric measures on the grain connectivity that help explaining the onset of shear band \citep{tordesillas2011discovering, tordesillas2010force}, coherent vortex structure \citep{williams1997coherent} and post-bifurcation behaviors in granular materials \citep{sun2013multiscale, wang2015anisotropy, liu2018coupled}. 
While these discoveries of new knowledge are indeed encouraging, one cannot  make use of them without investing significant efforts and time to derive, verify, and validate new constitutive laws that incorporate those new information. 
Hence the graph-theoretical approaches, although have found great promises on analyzing the granular assembles obtained from real or virtual experiments, have not yet made significant impacts on constitutive laws used for engineering applications. 
Our meta-modeling approach is capable of overcoming this bottleneck by efficiently automating some of these tasks currently undertaken by modelers. 
This second numerical experiment is used to demonstrate how the augmented intelligence can be used to incorporate the insights from observations into predictions without manually re-writing an existing constitutive law every time new information comes up. 

\begin{figure}[h!]\center
	\subfigure[Initial configuration of the "game board"]{
		\includegraphics[width=0.48\textwidth]{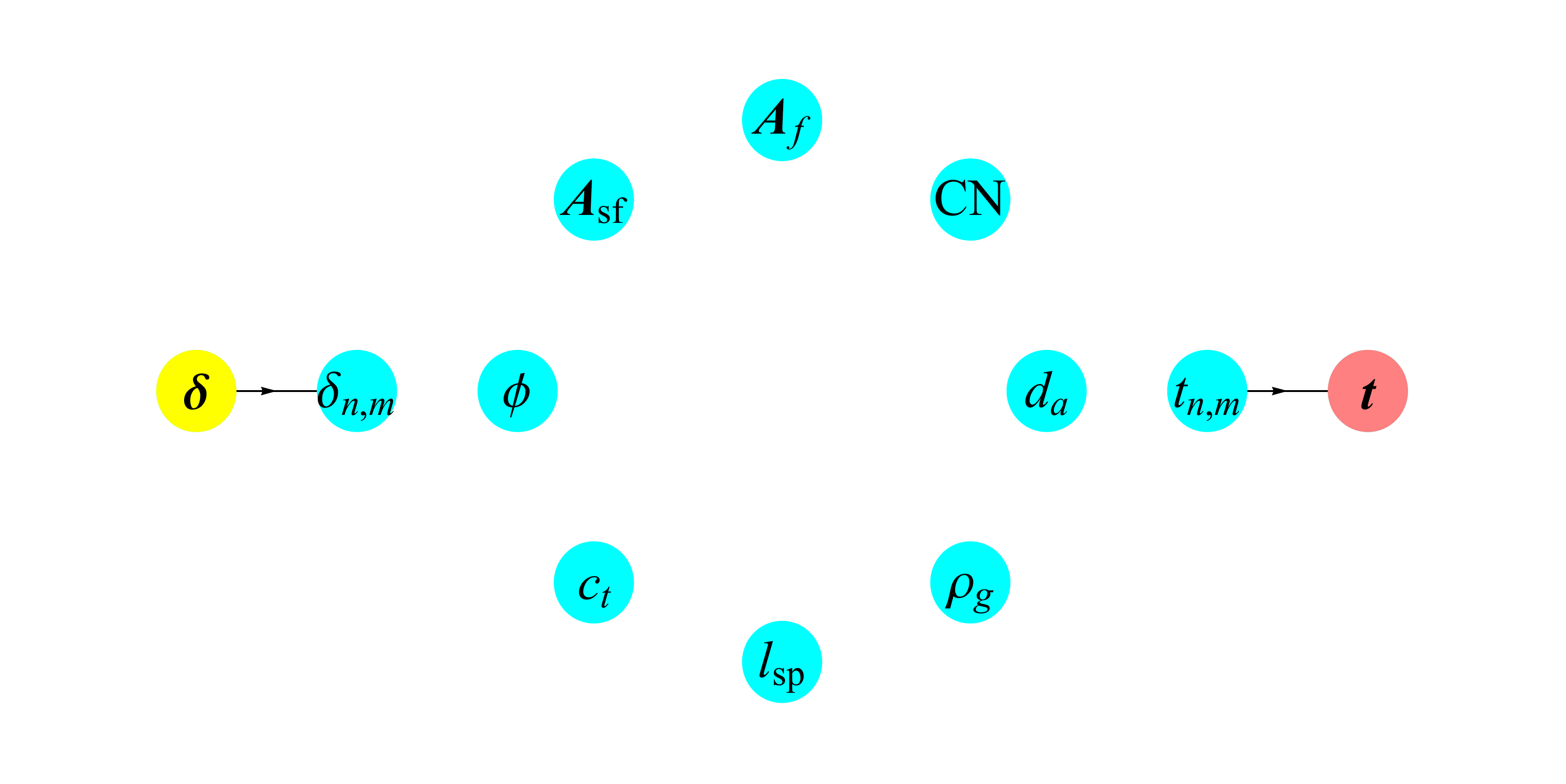}
	}
	\subfigure[All possible actions on the "game board"]{
		\includegraphics[width=0.48\textwidth]{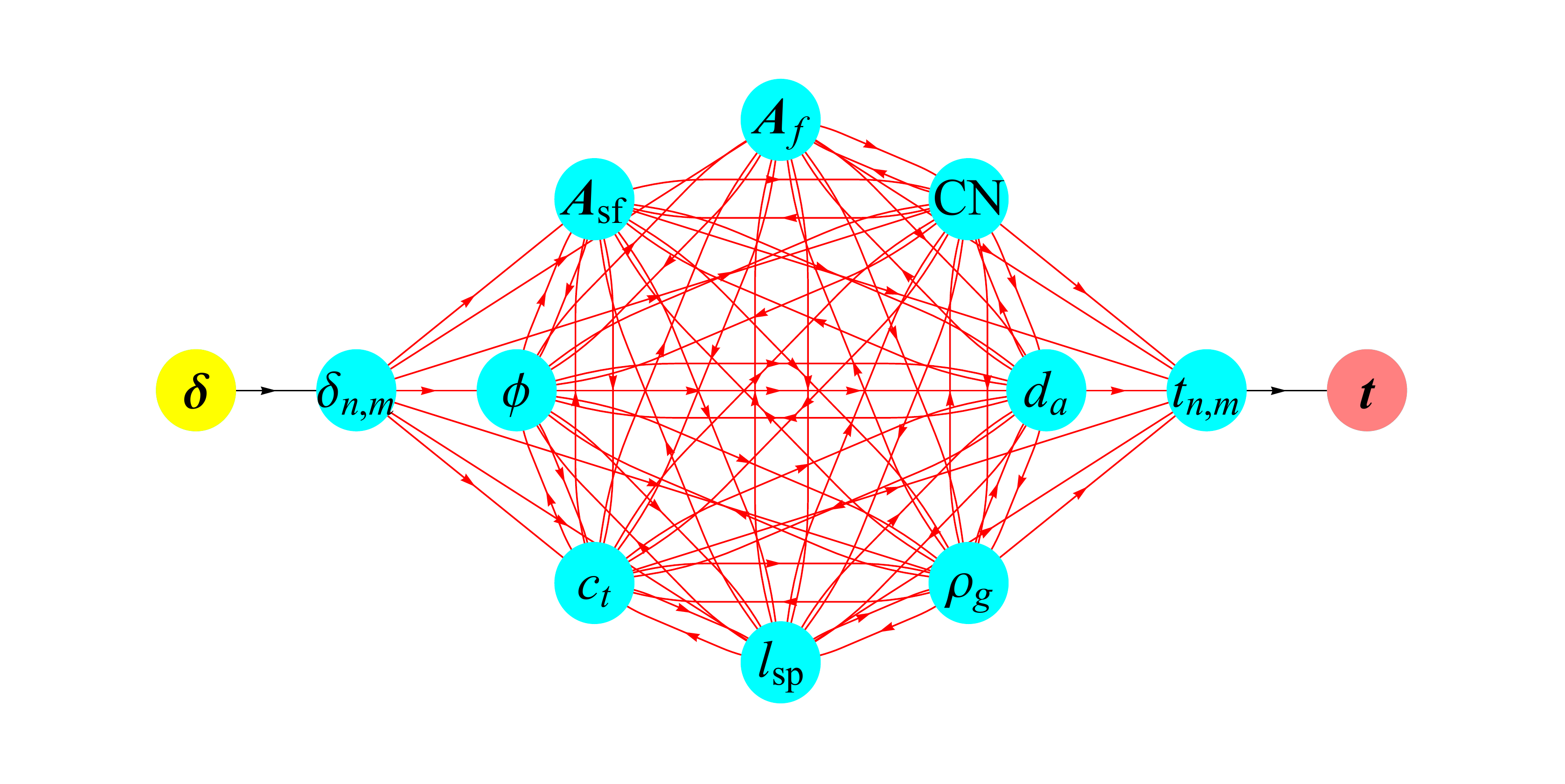}
	}
	\caption{A game of traction-separation model for a digraph involving the nodes $\{ \delta_{n,m},\ t_{n,m},\ \phi,\ CN,\ \tensor{A}_{f},\ \tensor{A}_{sf},\ d_{a},\ c_{t},\ l_{sp},\ \rho_{g} \}$ (detailed in Section \ref{sec:graph}). (a) the initial "board" on which the game is played. (b) All possible actions for picking the edges connecting the nodes are represented by the red arrows.}
	\label{fig:model_game2_board}
\end{figure}

This example is an extension of the first numerical experiment, in which more microstructual information are considered, including the fabric of strong interactions $A_{sf}$ and four measures of grain connectivities $d_{a},\ c_{t},\ l_{sp},\ \rho_{g}$. 
The task of identifying their roles in constitutive models for granular materials is now simply recast as defining a new game with augmented vertex set in digraph and extended action space. 
The "game board" and all possible actions for this new game are shown in Figure \ref{fig:model_game2_board}. 
The dimension of the action space increases from 13 to 71. 
A particular game rule is added to test the flexibility of the DRL algorithm in handling different types of game constraints: the strong fabric tensor $A_{sf}$ and the fabric tensor $A_f$, since both are geometric measures of inter-particles forces, are mutually exclusive in the final digraphs of constitutive models. 
The number of possible game states increases from 3200 to over 400000. 
The number of complete and admissible directed graph configurations increases from 591 to over 20000. 
The score definition is the same as Equation (\ref{eq:game1_score}). 
The meta-modeling algorithm try to learn the optimal ways to incorporate the microstructual information to make better predictions only from the training database of 50 loading cases, while the gained knowledge is validated on the test database of another 150 loading cases. 
The parameters for the DRL meta-modeling algorithm are set as: 10 iterations of "exploration and exploitation", 1 iteration of "competitive gameplay", 30 self-plays in each iteration, and 30 MCTS simulations in each play step. 

\begin{figure}[h!]\center
	\subfigure[Violin plots of the density distribution of model scores in each DRL iteration]{
		\includegraphics[width=0.45\textwidth]{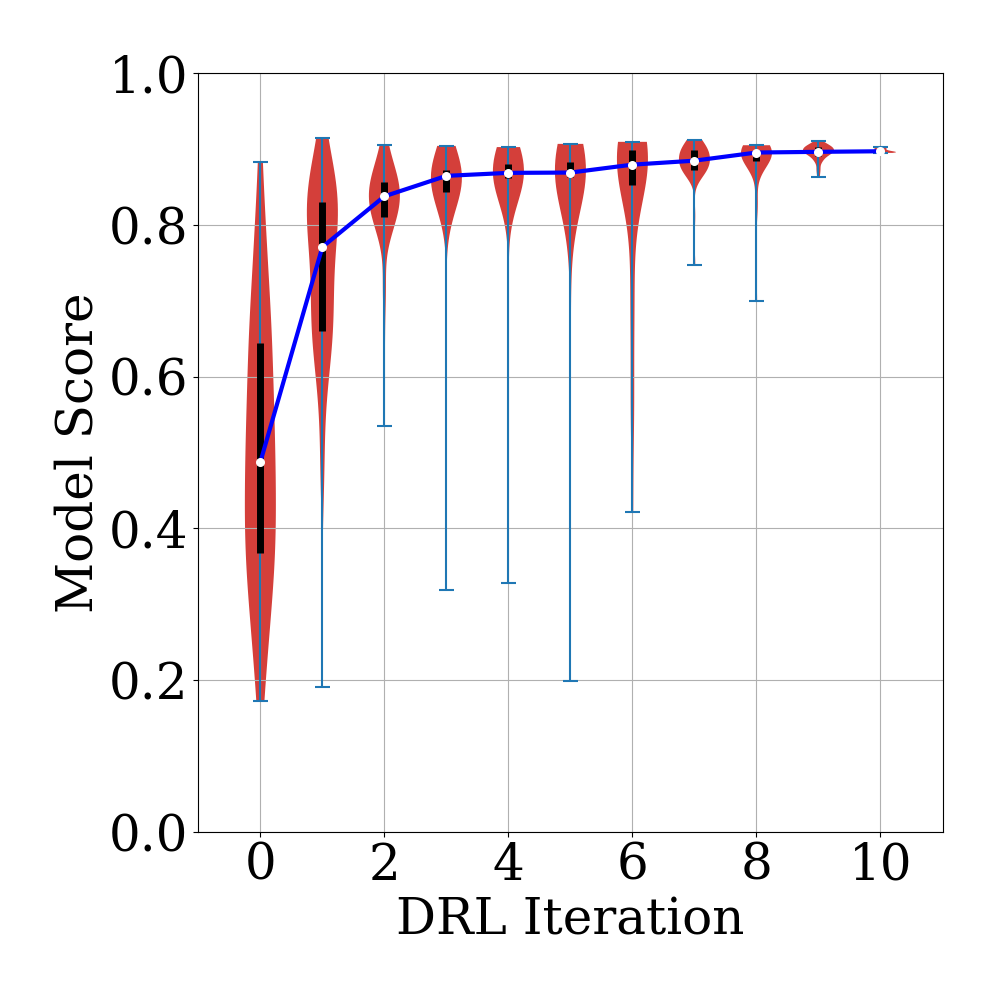}
	}
	\subfigure[Mean value and $\pm$ standard deviation of model score in each DRL iteration]{
		\includegraphics[width=0.45\textwidth]{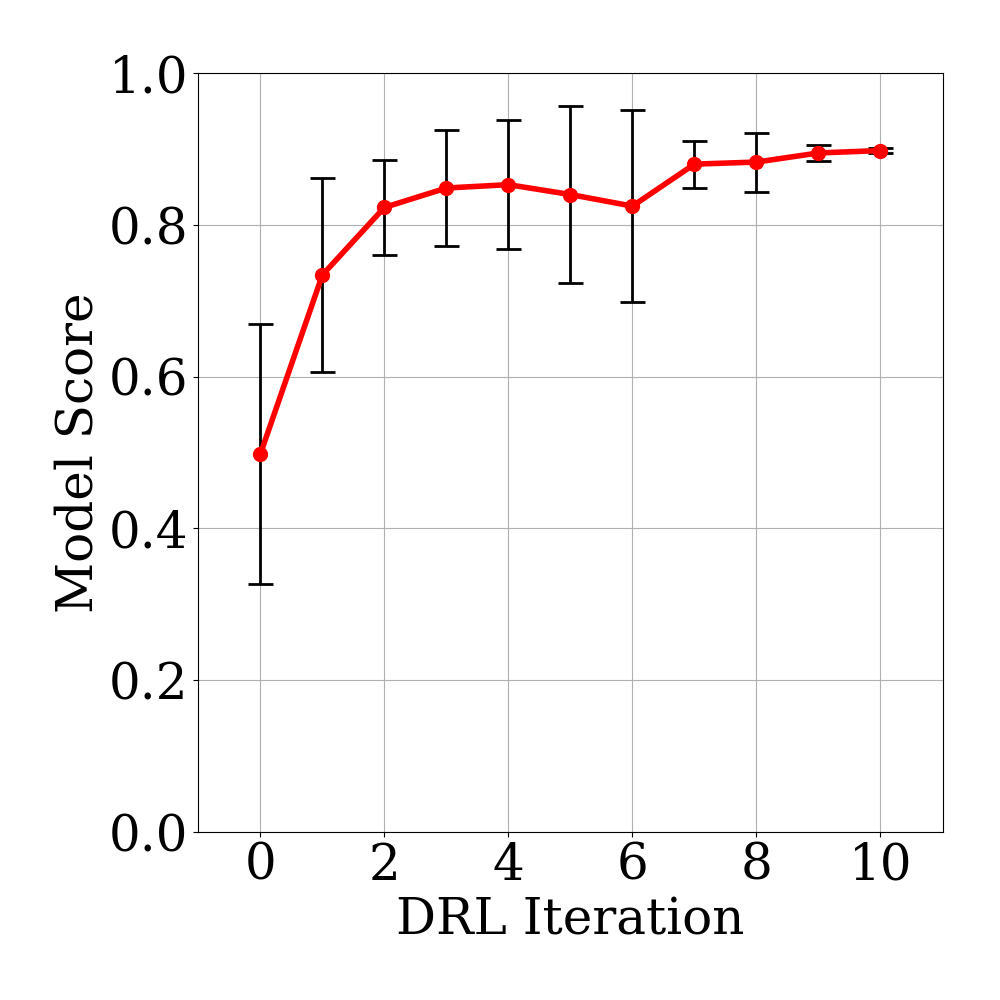}
	}
	\caption{Statistics of the model scores in deep reinforcement learning iterations from 5 separate runs of the DRL procedure for Numerical Experiment 2. Each DRL procedure contains ten iterations 0-9 of "exploration and exploitation" (by setting the temperature parameter $\tau=1.0$) and a final iteration 10 of "competitive gameplay" ($\tau=0.01$). Each iteration consists of 30 games. (a) Violin Plot of model scores played in each DRL iteration. The shade area represents the density distribution of scores. The white point represents the median. The thick black bar represents the interquartile range between 25\% quantile and 75\% quantile. The maximum and minimum scores played in each iteration are marked by horizontal lines. (b) Mean model score in each iteration and the error bars mark $\pm$ standard deviation.}
	\label{fig:game2_statistics}
\end{figure}

\begin{figure}[h!]\center
	\includegraphics[width=1.0\textwidth]{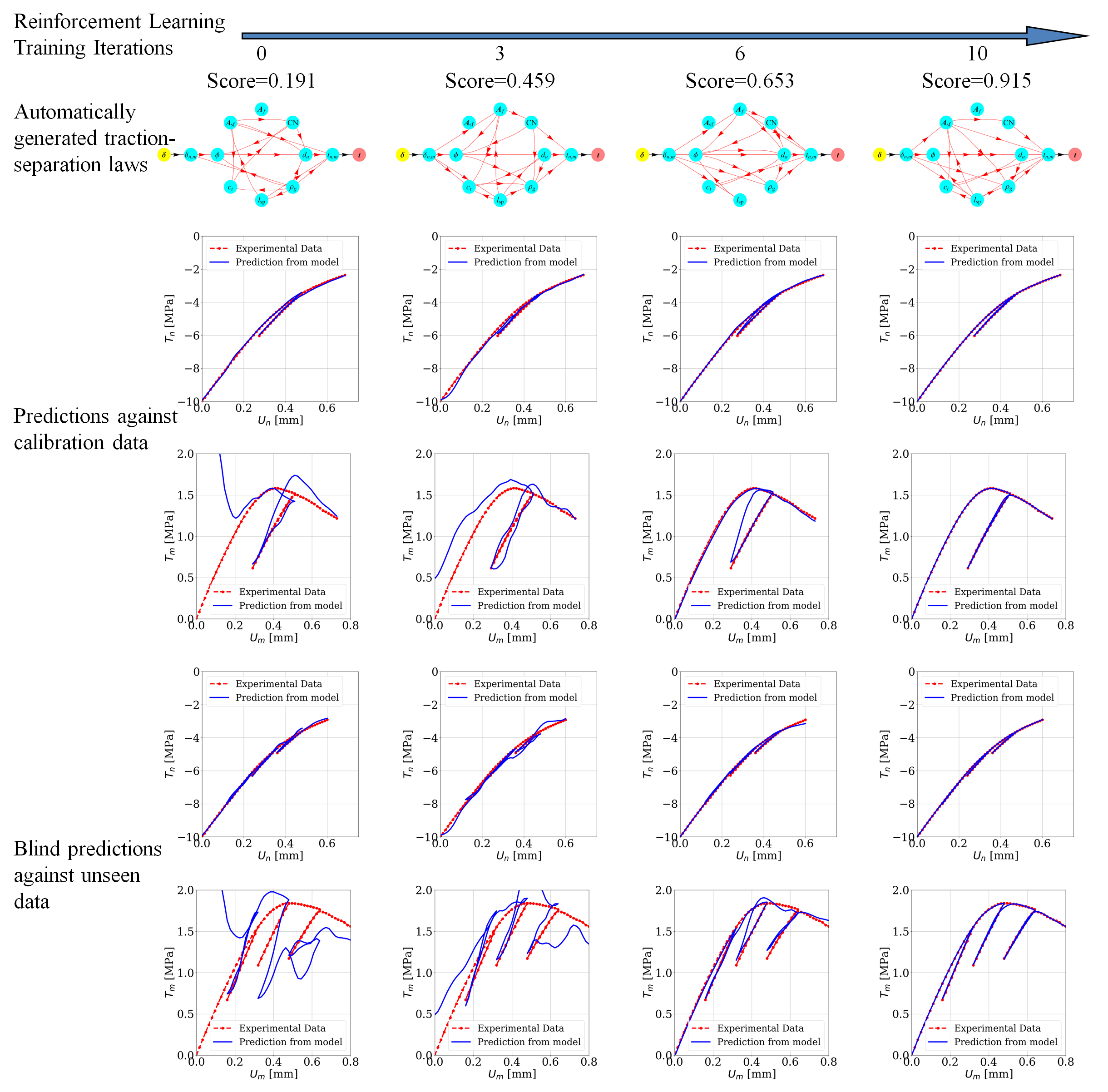}
	\caption{Knowledge of directed graphs of traction-separation models learned by deep reinforcement learning in Numerical Experiment 2.}
	\label{fig:model_drl_learn_game2}
\end{figure}

\begin{figure}[h!]\center
	\includegraphics[width=1.0\textwidth]{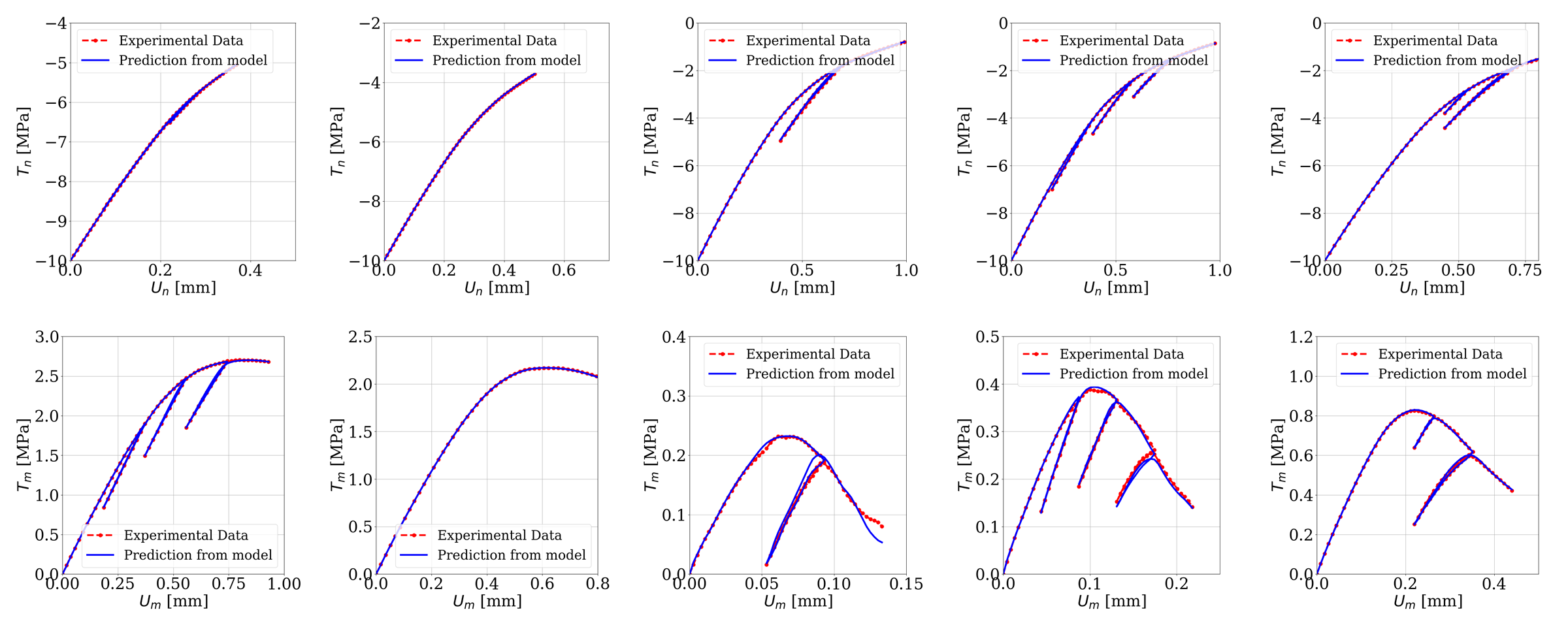}
	\caption{Five examples of blind predictions from the optimal digraph configuration (The 4th digraph in Figure \ref{fig:model_drl_learn_game2}) against unseen data among test database of 150 loading cases.}
	\label{fig:model_drl_predicts_game2}
\end{figure}

The statistics of the gameplay results from 5 separate runs of the DRL procedure are presented in Figure \ref{fig:game2_statistics}. 
We observe a very efficient improvement in generated traction-separation models, even though the number of legal game states in the new game has largely increased. 
Figure \ref{fig:model_drl_learn_game2} exhibits four representative digraph configurations developed during DRL iterations, as well as their prediction quality on calibration data and unseen data. 
It can be seen that the information flows in a constitutive model are of crucial importance. 
Although the first and the fourth graphs both incorporate the same types of microstructual information, the difference in the ways how these information are connected results in significant difference in model scores of 0.191 and 0.915, respectively. 
Moreover, the DRL algorithm develops the intelligence of selecting the strong fabric tensor $A_{sf}$ over the fabric tensor $A_f$ in order to further improve the prediction score of the model. 
Five blind prediction examples of the optimal digraph configuration (the 4th digraph in Figure \ref{fig:model_drl_learn_game2}) obtained in this game are presented in Figure \ref{fig:model_drl_predicts_game2}. 
Comparing to the numerical example 1 (Figure \ref{fig:model_drl_predicts}), the augmented knowledge of additional microstructural information in constitutive models lead to more accurate representations of granular materials. 

\subsection{Application: Multiscale bridging using DRL-generated traction-separation model in finite element modeling}
\begin{figure}[h!]\center
	\subfigure[Embedded interface of sinusoidal line]{
		\includegraphics[width=0.4\textwidth]{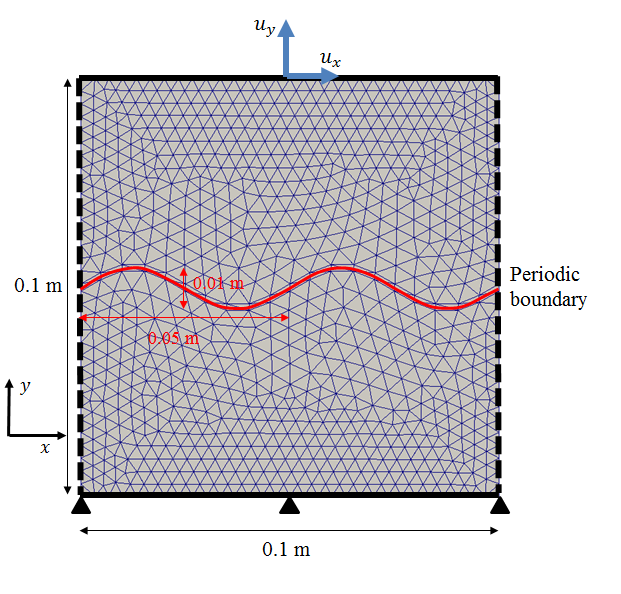}
	}
	\subfigure[Prescribed displacement path on the top edge]{
		\includegraphics[width=0.4\textwidth]{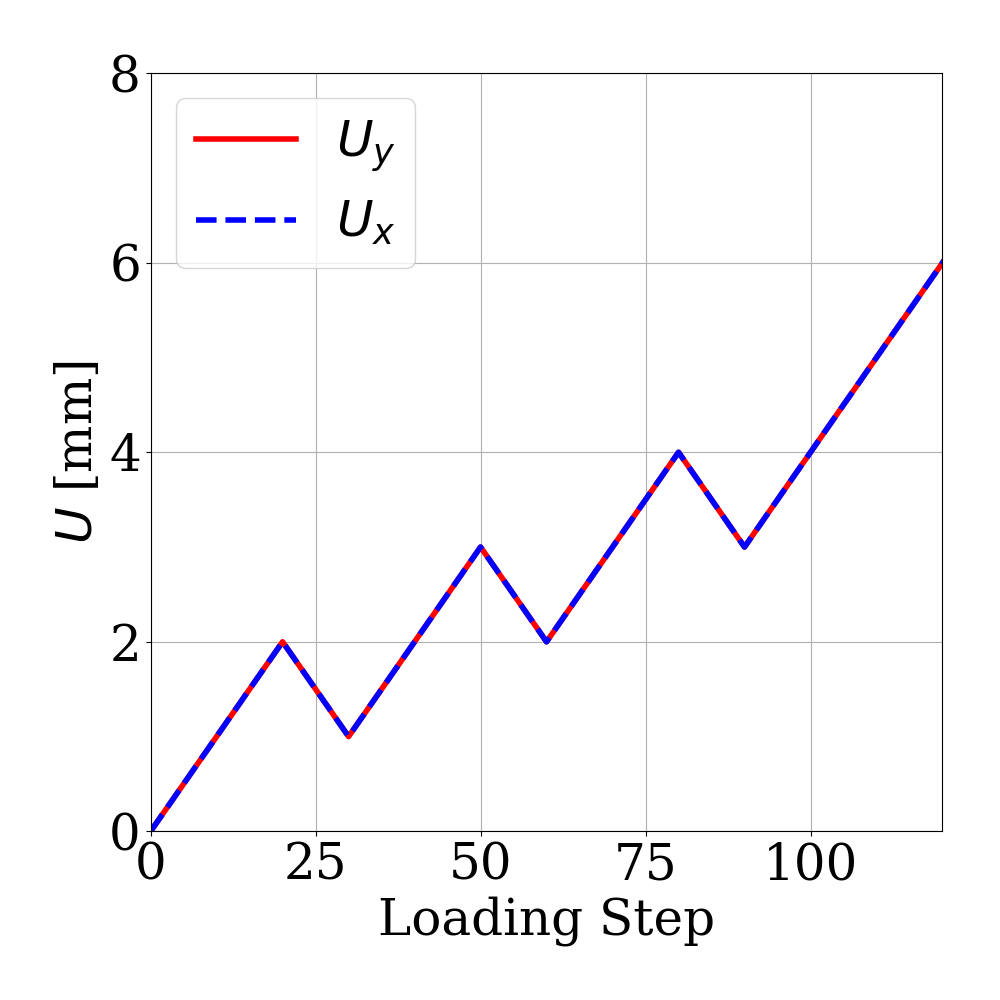}
	}
	\caption{Plane-strain mixed-mode tension-shear test on a granular specimen with pre-existing interface. Geometry, mesh and boundary conditions.}
	\label{fig:sinus_mesh}
\end{figure}

In this example, we demonstrate that the traction-separation model auto-generated by the proposed DRL meta-modeling can be applied in multiscale finite element simulations, as a surrogate model to replace the computationally expensive DEM RVEs. 
The example consists of a mixed-mode tension-shear test on a plane-strain granular specimen with embedded strong discontinuity. 
The sample is 0.1 m x 0.1 m in dimension. The bottom edge is fixed, while the top edge moves rigidly following a mixed-mode loading-unloading-reloading displacement path, as shown in Fig. \ref{fig:sinus_mesh}. 
The sample is assumed periodic in the horizontal direction, thus periodic displacement boundary condition is applied on the lateral edges. 
The geometry of the pre-existing interface in the center is a sinusoid with the spatial period of 0.05 m and amplitude of 0.005 m. 
The elements along the interface are enhanced by assumed strain formulation to embed the strong discontinuity \citep{oliver2002continuum}, while the other elements are regular bulk finite elements. 
The optimal traction-separation model found in the Numerical Example 2 is applied in the interface. 
The bulk material is assumed isotropic linear elastic and the parameters are homogenized from the DEM RVE for generation of data (Young's modulus $E=300$ MPa, Poisson's ratio $\nu=0.24$). 
The specimen is initially under isotropic pressure of $10$ MPa, the same as the DEM RVEs, and the lateral pressure remains constant during the loading steps.

\begin{figure}[h!]\center
	\subfigure[Load step 50]{
		\includegraphics[width=0.3\textwidth]{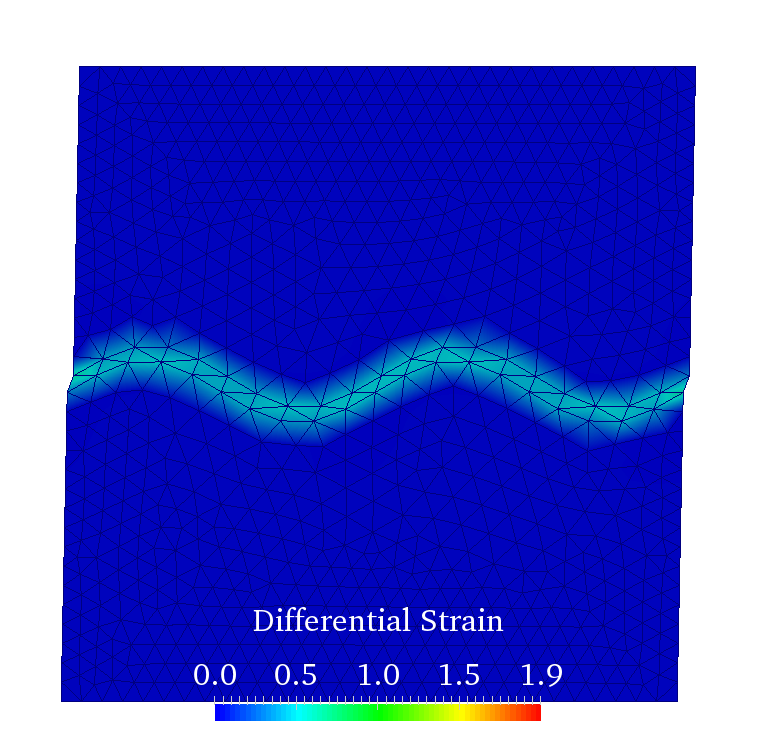}
	}
	\subfigure[Load step 80]{
		\includegraphics[width=0.3\textwidth]{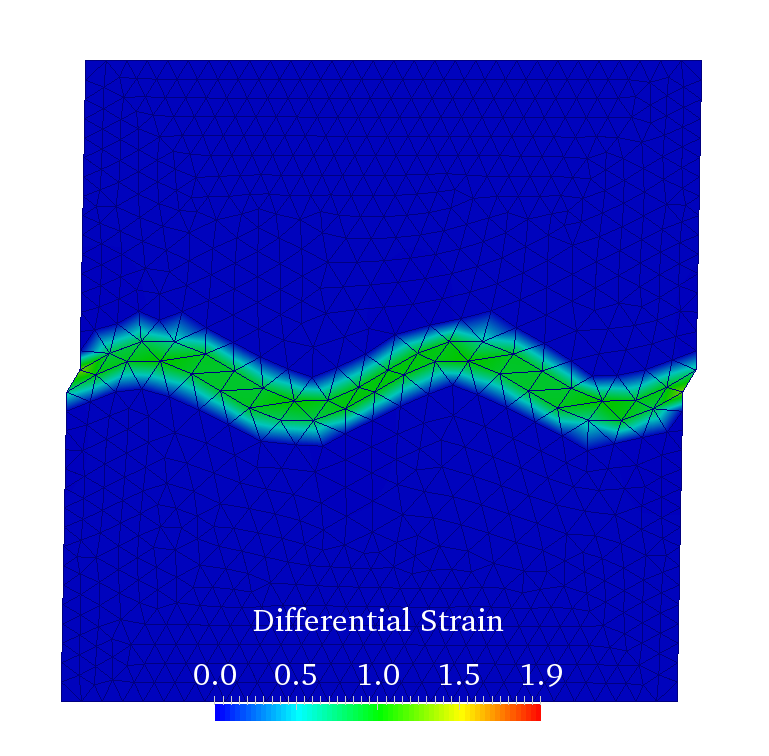}
	}
	\subfigure[Load step 110]{
		\includegraphics[width=0.3\textwidth]{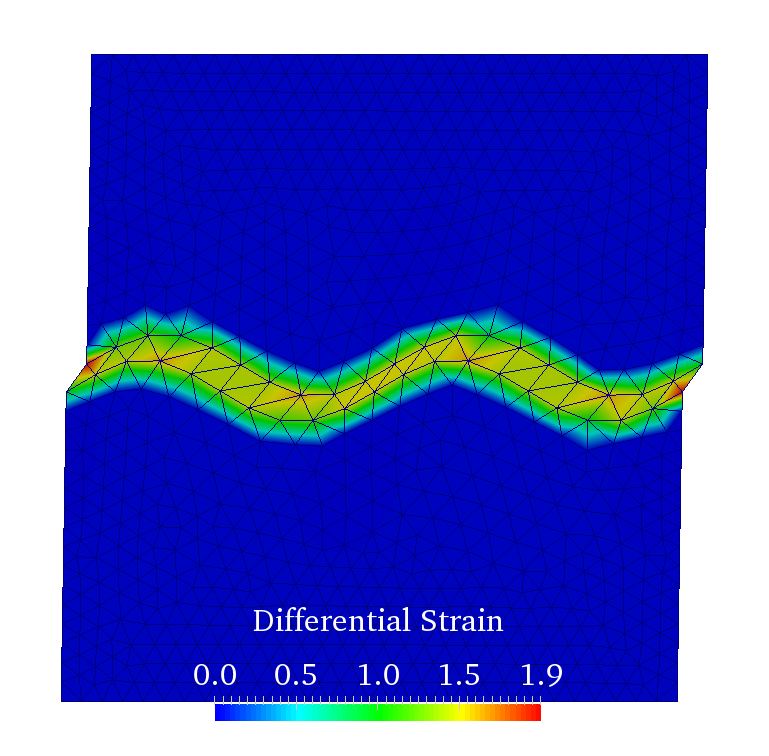}
	}
	\subfigure[Load step 50]{
		\includegraphics[width=0.3\textwidth]{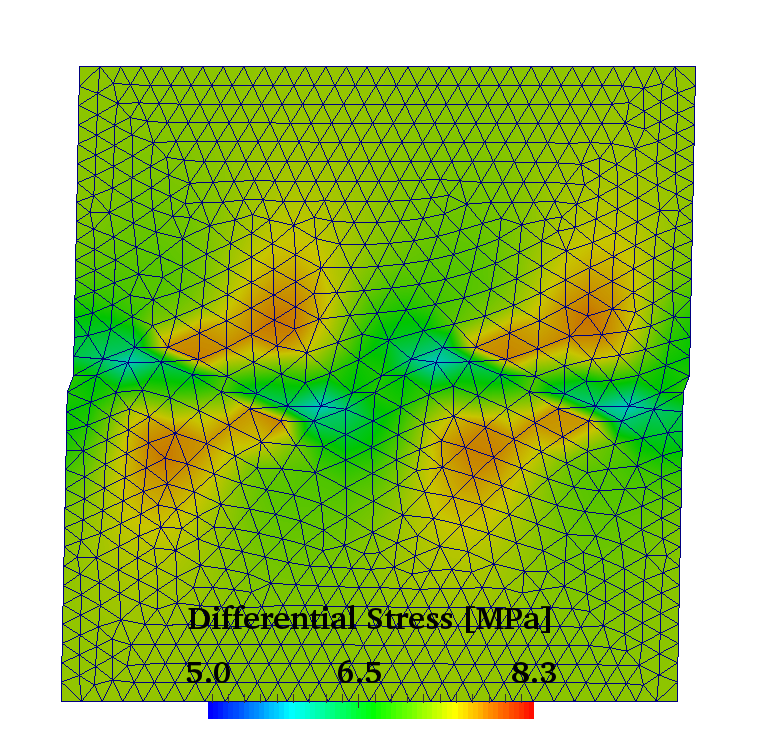}
	}
	\subfigure[Load step 80]{
		\includegraphics[width=0.3\textwidth]{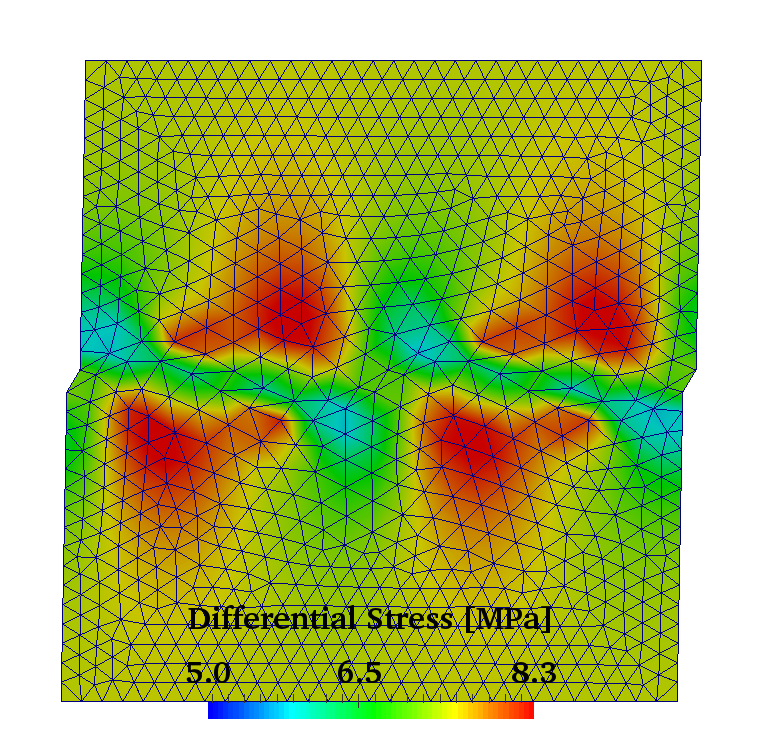}
	}
	\subfigure[Load step 110]{
		\includegraphics[width=0.3\textwidth]{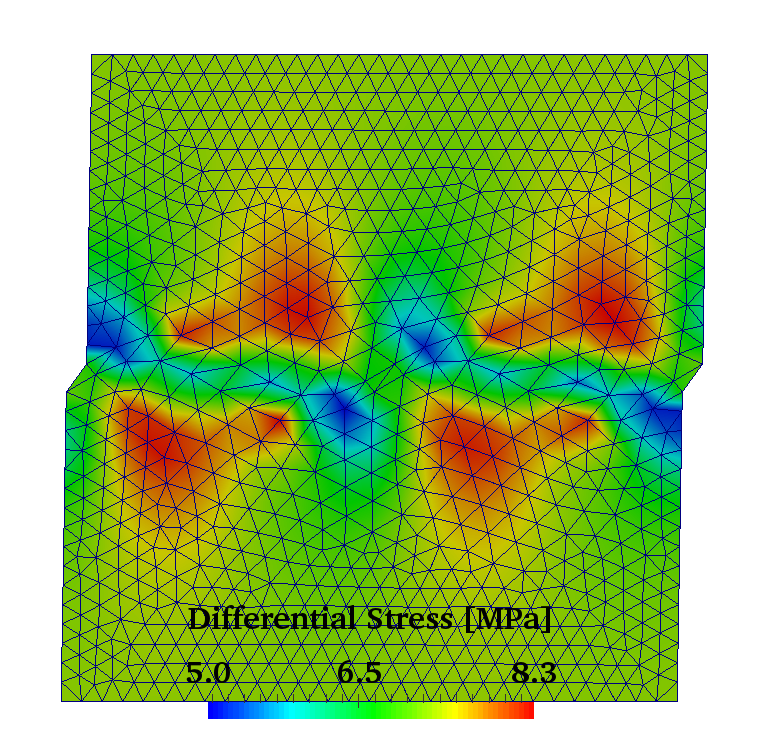}
	}
	\caption{Differential stress field ($\sigma_1-\sigma_3$, where $\sigma_1$ is the greatest principal value and $\sigma_3$ is the least principal value of the Cauchy stress tensor $\tensor{\sigma}$) and differential strain field ($\varepsilon_1-\varepsilon_3$, where $\varepsilon_1$ is the greatest principal value and $\varepsilon_3$ is the least principal value of the strain tensor $\tensor{\varepsilon}$) at loading steps 50, 80 and 110 in the mixed-mode tension-shear test.}
	\label{fig:sinus_test_diffstressstrain}
\end{figure}

\begin{figure}[h!]\center
	\subfigure[Embedded interface of sinusoidal line]{
		\includegraphics[width=0.4\textwidth]{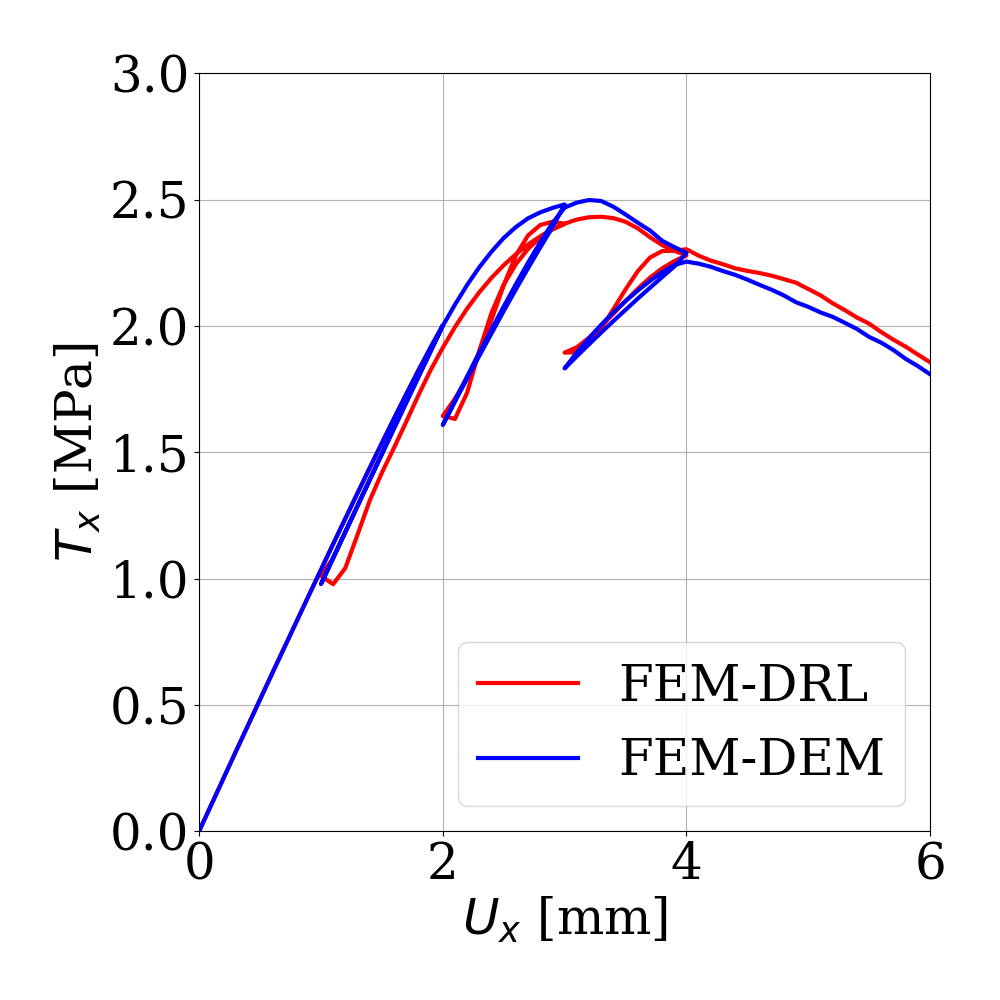}
	}
	\subfigure[Prescribed displacement path on the top edge]{
		\includegraphics[width=0.4\textwidth]{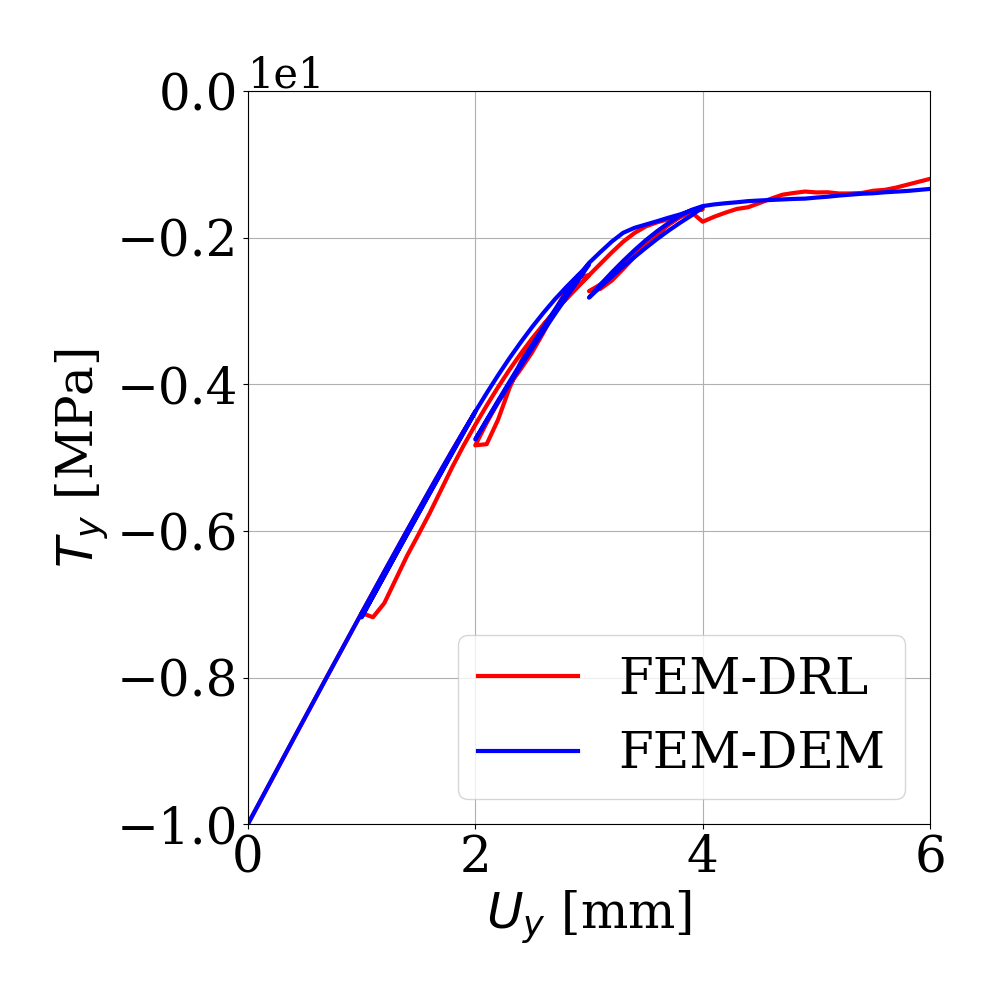}
	}
	\caption{Comparison of FEM-DRL and FEM-DEM multiscale simulations on global traction-displacement relation in normal ($U_y-T_y$) and tangential ($U_x-T_x$) directions on the top edge of the specimen.}
	\label{fig:sinus_test_curve}
\end{figure}

The differential stress and strain fields computed by FEM-DRL multiscale approach at selected loading steps are shown in Fig. \ref{fig:sinus_test_diffstressstrain}.
The shear strain localizes in the embedded strong discontinuity in both specimens. 
The global traction-displacement curves in normal ($U_y-T_y$) and shear ($U_x-T_x$) directions are compared for both FEM-DRL and FEM-DEM simulations in Fig. \ref{fig:sinus_test_curve}. 
Results show great agreement on the mechanical behavior of the specimen. 
Hence the optimal traction-separation model automatically developed by the DRL meta-modeling approach can be used as a surrogate model to microscale DEM RVEs in multiscale FEM simulations.

\section{Conclusion}
This paper presents a meta-modeling approach in which we attempt to generate traction-separation laws not through explicitly writing a 
particular model but to provide the computer with modeling options such that it can explore on its own through self-practicing. 
Unlike previous deep-learning models that leverage supervised learning techniques to train neural networks that makes black-box predictions, this new approach focuses on reinforcement learning technique to discover hidden relationships among data and therefore make modeling decisions to emulate the process of writing constitutive models by human. 
Given the rules (frame indifference, thermodynamic laws, balance principles), we introduce an agenda-based approach where the DRL technique is used to find the optimal way to generate a forward
prediction. As demonstrated in our numerical experiments, this approach can be regarded as a generalization of the previous models where 
neural network predictions may still embed in part of the predictions but are not necessarily completely replacing all components in the conventional models. This flexibility is the key for us to exploit the computer to make \textit{repeated} trial-and-errors and improve from experiments over time to generate the best outcomes, instead of spending significant human time to explore through trial-and-errors. 
The idea of inventing the metal-modeling game could be significant in the sense that it frees modelers from focusing on curve-fitting a physical process.  Instead, as future improvements on the models can be made by expanding the action space or simply leveraging the power of computer to improve the models over time, this allows us unprecedented luxury to place our focuses on finding best cause of actions that lead to the most predictive model. 
In addition, this meta-modeling approach also provides the following unique benefits against the conventional hand-crafting approach and black-box neural network models. 
\begin{enumerate}
	\item Since the machine learning procedure is automated, models intended for different purposes or designed to fulfill different demands (speed, accuracy, robustness) can be automatically generated and improved over time through self-plays in the model-creation game. 
	\item Since the validation procedure is introduced as the reward mechanism for the agent to find 
	the optimal models available, the resultant models are always validated at the end of the games. 
	\item By recasting constitutive models as directed graphs, previous models established by domain experts can be easily embedded in the proposed framework to expand action spaces efficiently and shorten the training time. 
	\item The metal-modeling approach is generic and reusable, which means that it can handle different situations with different data, objective functions and rules set by human without going through additional derivation, implementation, material parameter identification and validation. Hence it does not require any debugging once the the game is implemented correctly.  
\end{enumerate}

There are also limitations of the current approaches. For instance, the demands for data could be higher than
conventional modeling approach, particularly when more sophisticated and rigorous validation metric is used to assign model score and game reward. 
The model has not yet introduced any technique to handle noise, nor does it consider any mechanism to consider uncertainty. 
While these topics are of great importance, the corresponding research activities are beyond the scope of this study and will be 
considered in the future, should opportunities come. 

\section*{Acknowledgments}
The corresponding author's work is supported by the Earth Materials and Processes
program from the US Army Research Office under grant contract 
W911NF-15-1-0442 and W911NF-15-1-0581, 
the Dynamic Materials and Interactions Program from the Air Force Office of Scientific Research
under grant contract FA9550-17-1-0169, 
the nuclear energy university program from department of energy under grant contract DE-NE0008534 as well as
the Mechanics of Material program at National Science Foundation under grant 
contract CMMI-1462760. 
Meanwhile, the first author is supported by US Army Research Office under grant contract 
W911NF-15-1-0442 and W911NF-15-1-0581, and the 2018 Interdisciplinary Research Seed funding from 
Columbia University. 
These supports are gratefully 
acknowledged. 
The views and conclusions contained in this document are those of the authors, 
and should not be interpreted as representing the official policies, either expressed or implied, 
of the sponsors, including the Army Research Laboratory or the U.S. Government. 
The U.S. Government is authorized to reproduce and distribute reprints for 
Government purposes notwithstanding any copyright notation herein.

\section*{Appendix A: Generation of synthetic data from discrete element modeling (DEM)}
The data for calibration and evaluation of prediction accuracy of the deep-reinforcement-learned traction-separation models are generated by numerical simulations on a representative volume element (RVE) representing the granular materials on a frictional surface. 
The open-source software YADE for DEM is used \citep{vsmilauer2010yade}. 
The discrete element particles in the RVE have radii between $1 \pm 0.3$ mm with uniform distribution. 
The RVE has the height of $20$ mm in the normal direction of the frictional surface and is initially consolidated to isotropic pressure of 10 MPa. 
The Cundall's elastic-frictional contact model (\citep{cundall1979discrete}) is used for the inter-particle constitutive law. 
The material parameters are: interparticle elastic modulus $E_{eq}=1$ GPa, ratio between shear and normal stiffness $k_s/k_n=0.3$, frictional angle $\varphi=$ \ang{30}, density $\rho=2600$ $kg/m^3$, Cundall damping coefficient $\alpha_{damp}=0.2$. 

\begin{figure}[h!]\center
	\subfigure[RVE of frictional surface]{
		\includegraphics[width=0.23\textwidth]{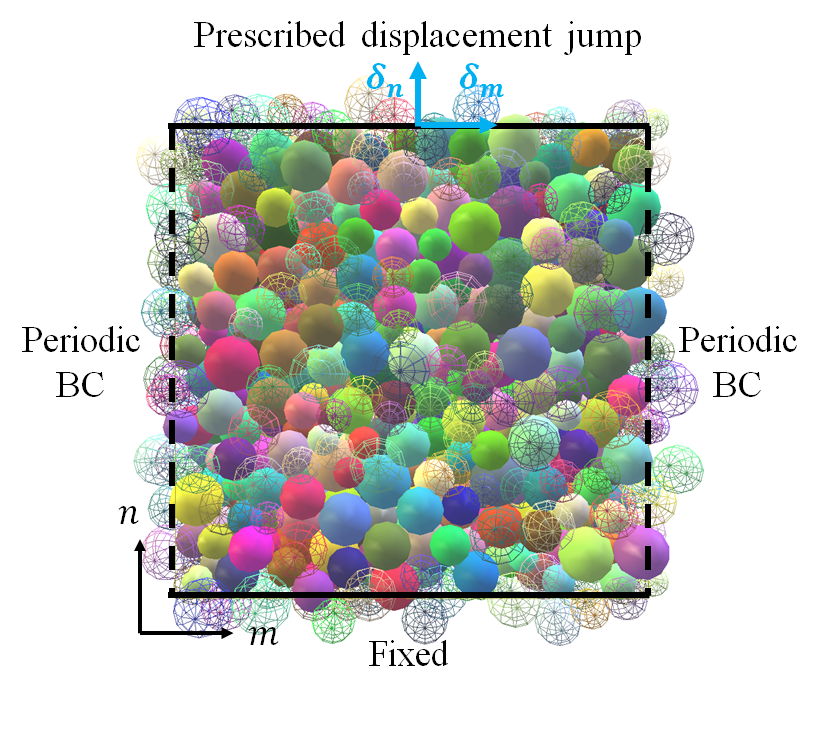}
		\label{fig:DEMRVE_load_path_1}
	}
	\subfigure[Example loading path 1]{
		\includegraphics[width=0.23\textwidth]{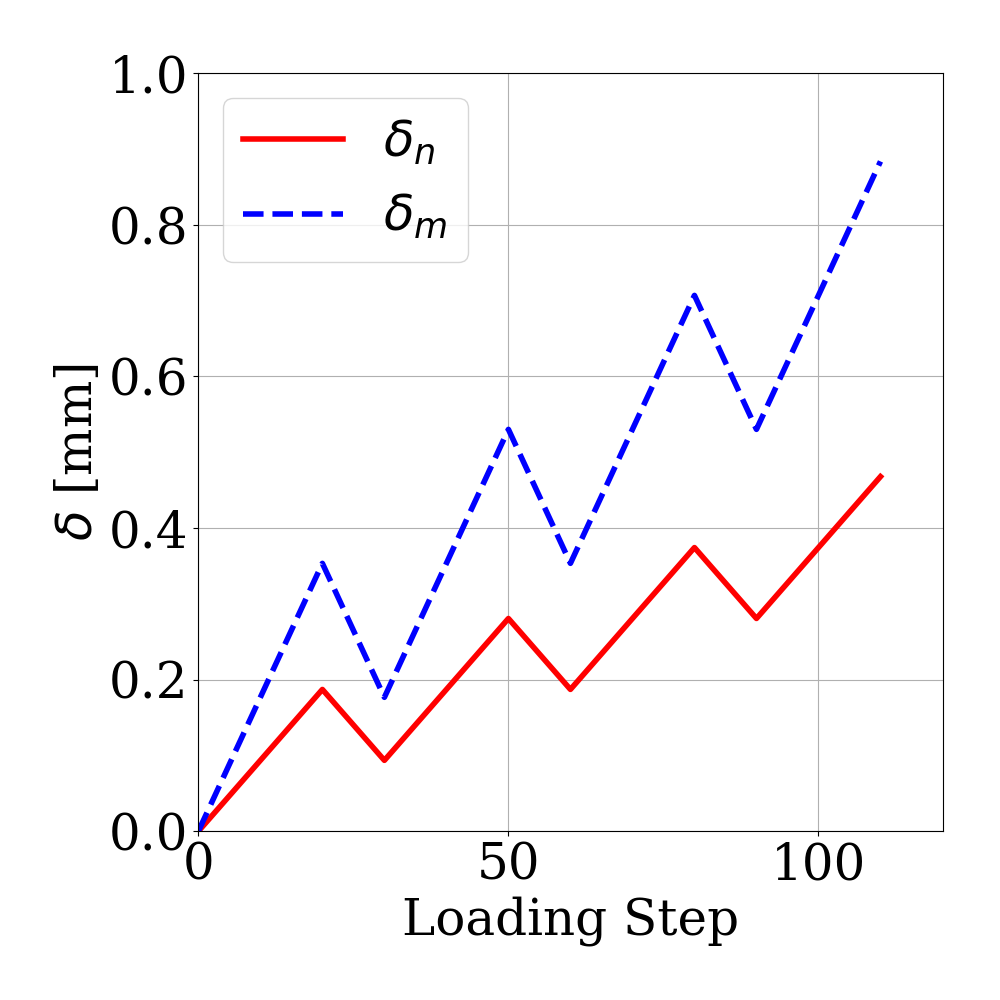}
		\label{fig:DEMRVE_load_path_2}
	}
	\subfigure[Example loading path 2]{
		\includegraphics[width=0.23\textwidth]{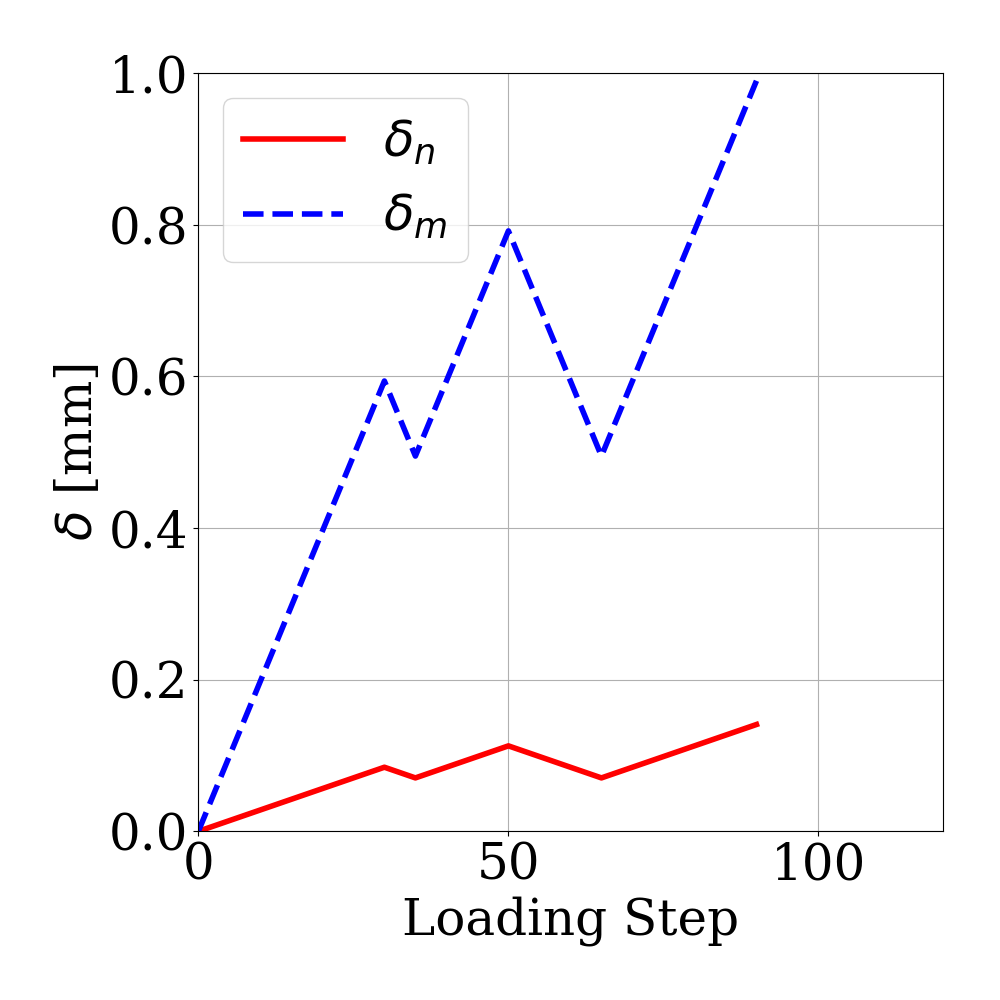}
		\label{fig:DEMRVE_load_path_3}
	}
	\subfigure[Example loading path 3]{
		\includegraphics[width=0.23\textwidth]{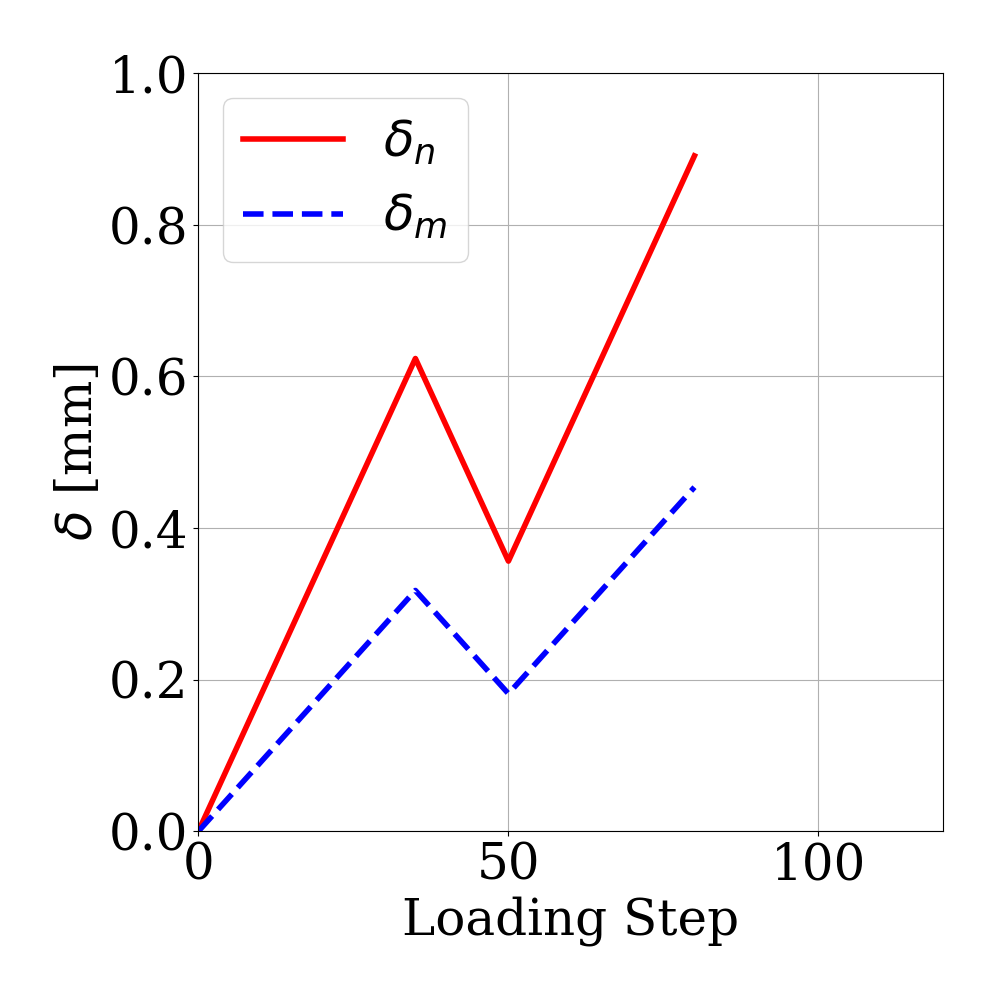}
		\label{fig:DEMRVE_load_path_4}
	}
	\caption{Representative volume element of a frictional surface having normal $\vec{n}$ and tangential $\vec{m}$ directions. Three examples of different loading-unloading-reloading paths among 200 numerical experiments for the generation of database are shown.}
	\label{fig:DEMRVE_load_path}
\end{figure}

\begin{figure}[h!]\center
	\subfigure[Example loading path 1]{
		\includegraphics[width=0.31\textwidth]{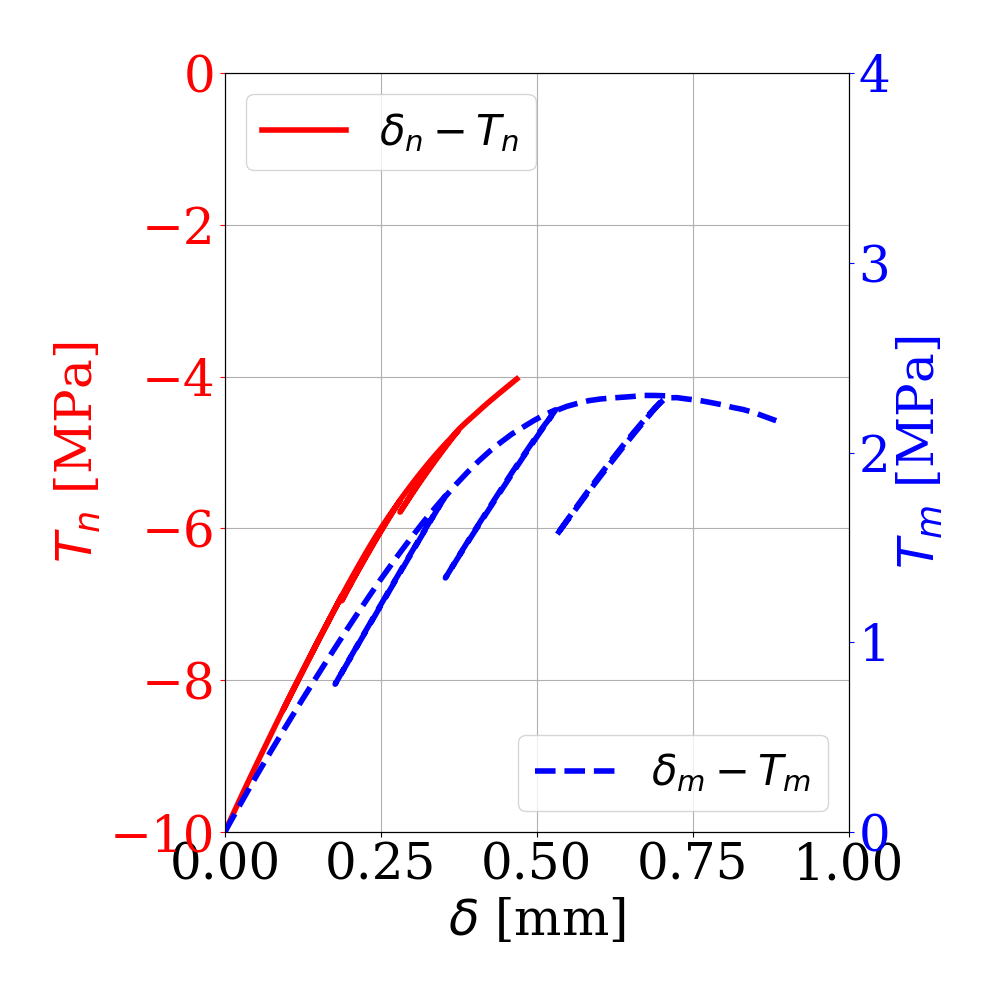}
		\label{fig:DEMRVE_UT_1}
	}
	\subfigure[Example loading path 2]{
		\includegraphics[width=0.31\textwidth]{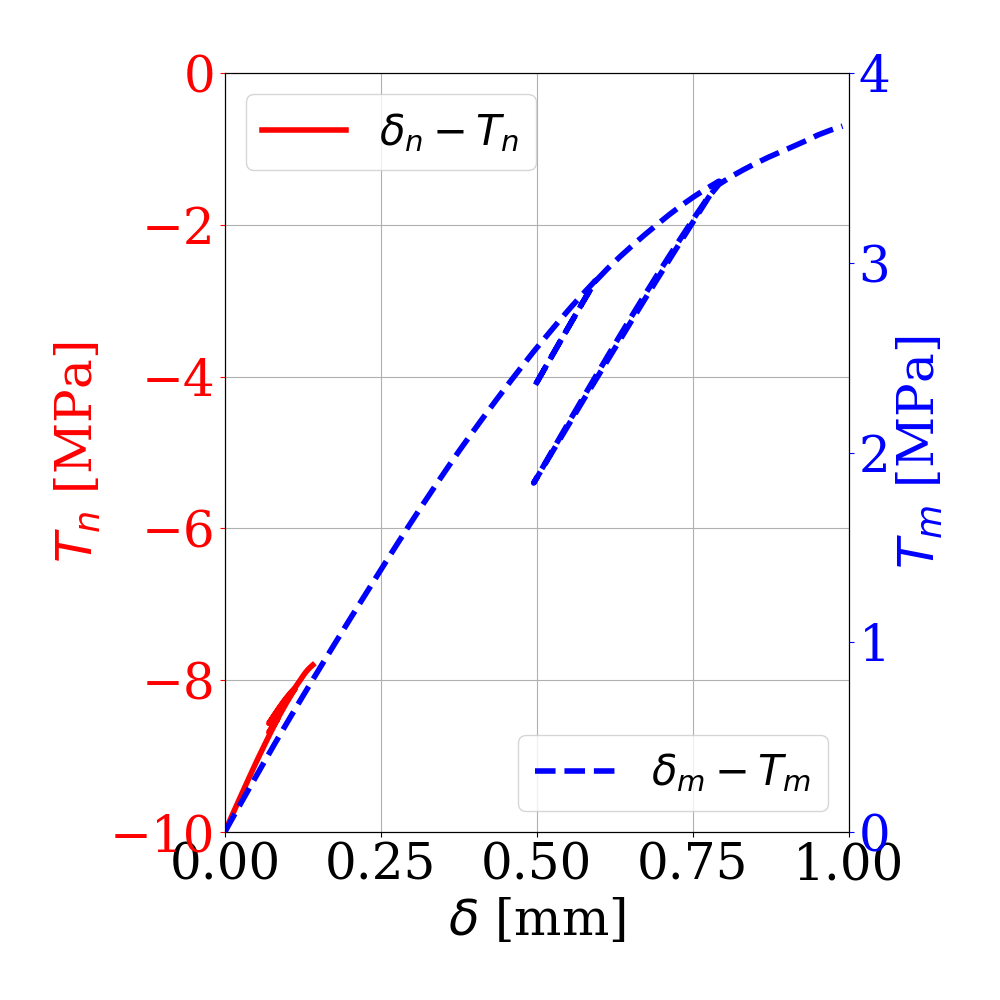}
		\label{fig:DEMRVE_UT_2}
	}
	\subfigure[Example loading path 3]{
		\includegraphics[width=0.31\textwidth]{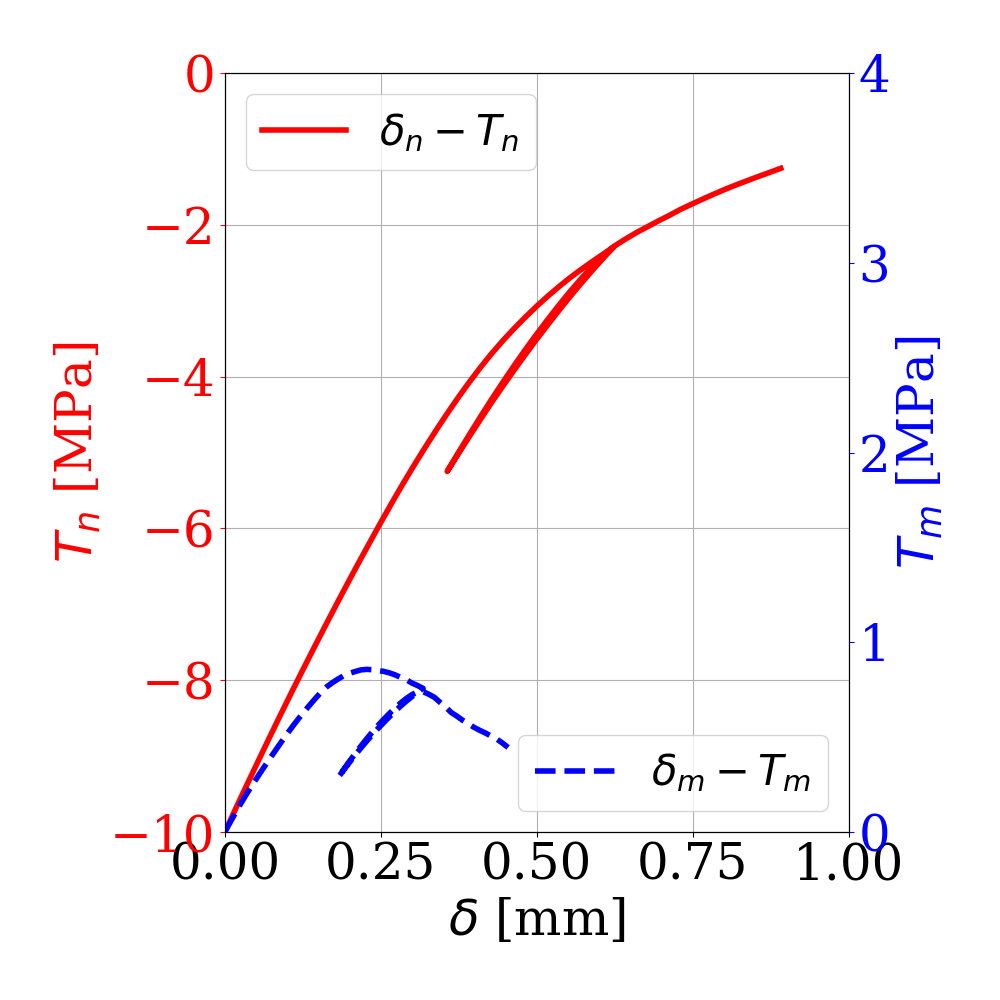}
		\label{fig:DEMRVE_UT_3}
	}
	\caption{Examples of traction-separation curves corresponding to the three loading paths in Figure \ref{fig:DEMRVE_load_path} among 200 numerical simulations. The normal traction $T_n$ is plotted against the normal displacement jump $\delta_n$. The tangential traction $T_m$ is plotted against the tangential displacement jump $\delta_m$.}
	\label{fig:DEMRVE_UT}
\end{figure}

\begin{figure}[h!]\center
	\subfigure[Coordination number]{
		\includegraphics[width=0.31\textwidth]{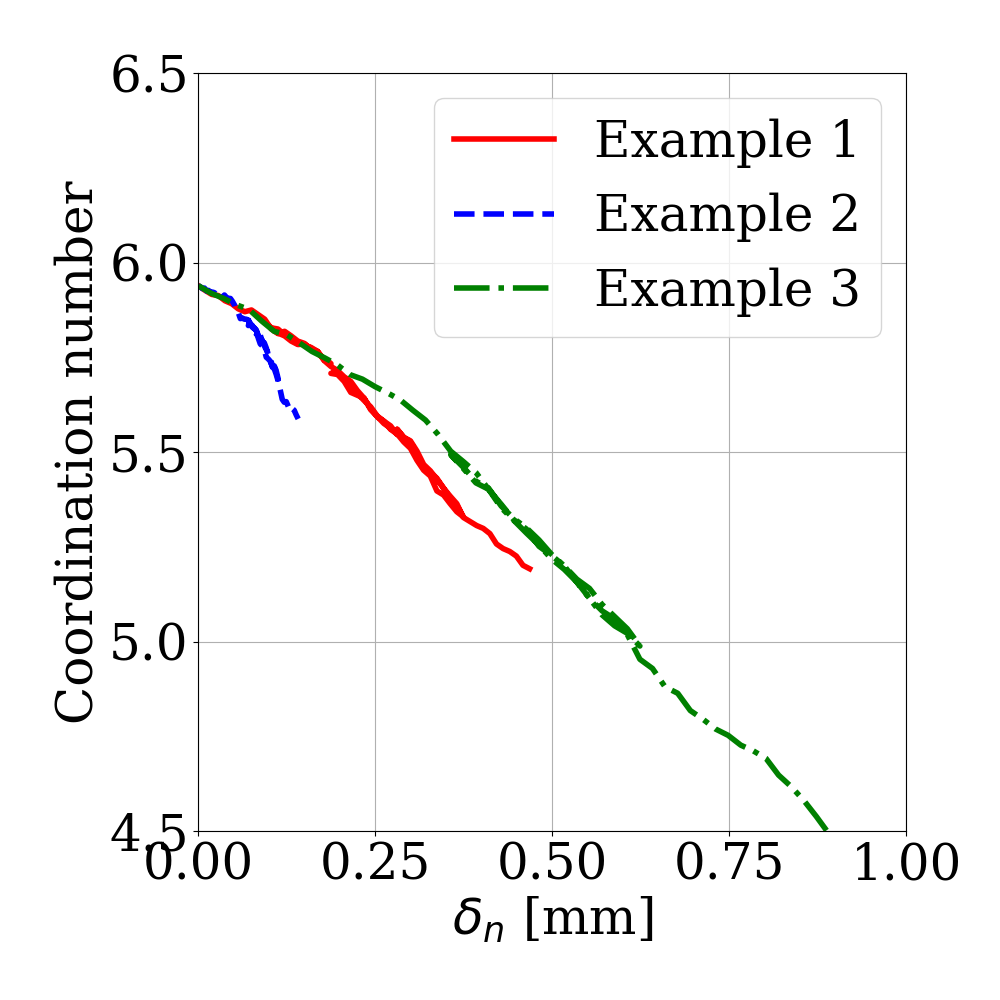}
	}
	\subfigure[Average Shortest Path Length]{
		\includegraphics[width=0.31\textwidth]{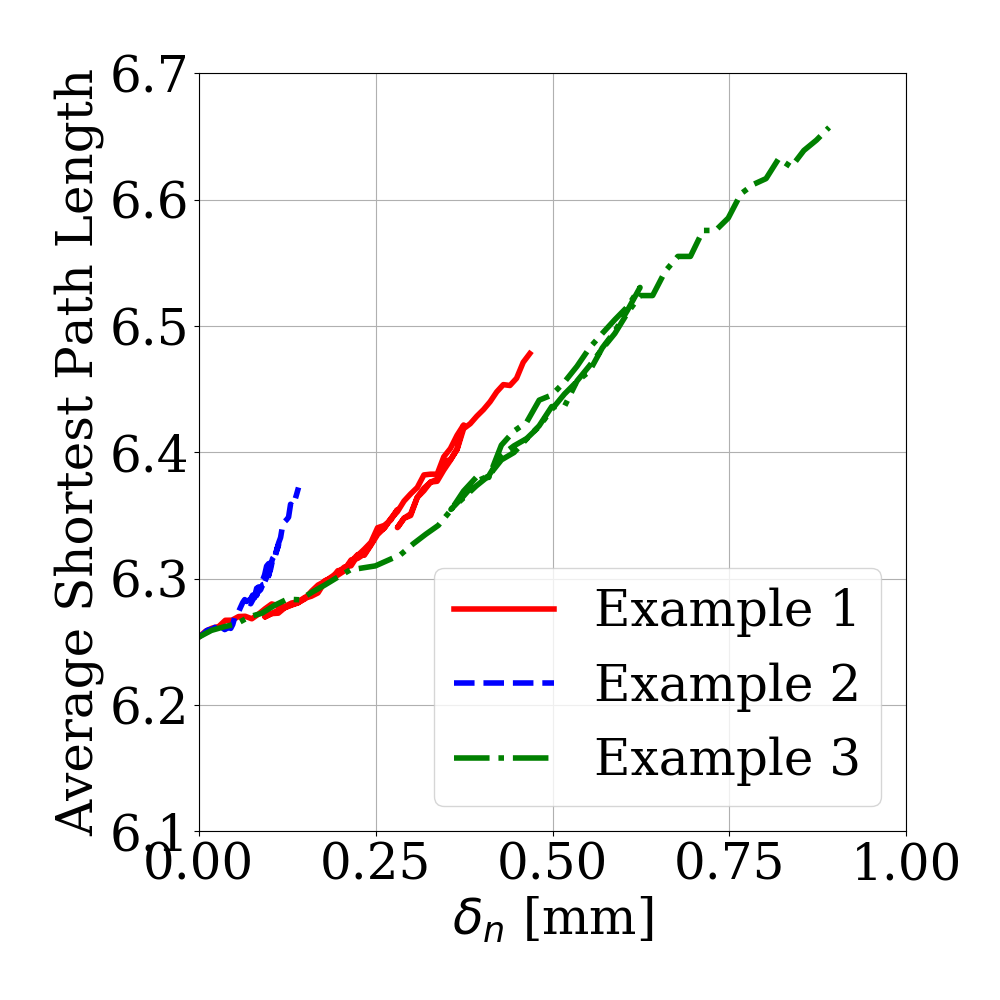}
	}
	\subfigure[Graph density]{
		\includegraphics[width=0.31\textwidth]{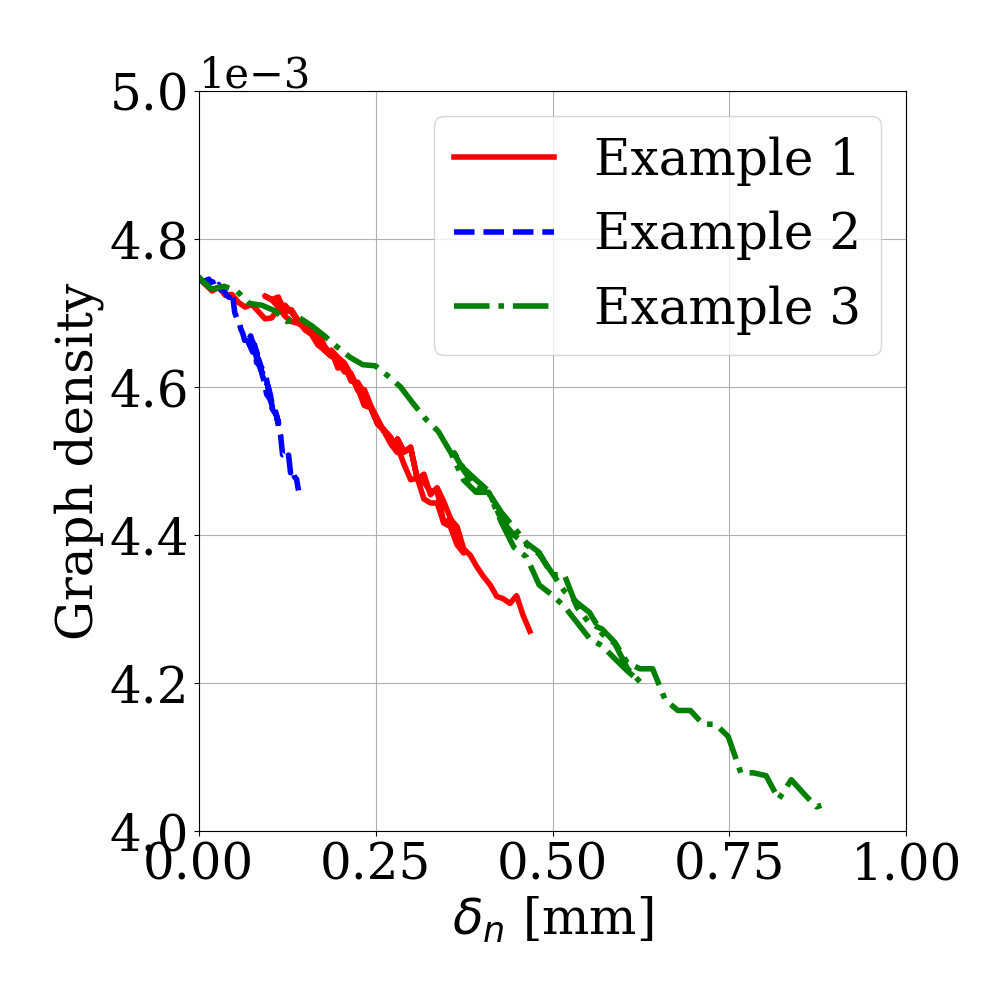}
	}
	\caption{Examples of coordination number, average shortest path length and graph density for particle interactions corresponding to the three loading paths in Figure \ref{fig:DEMRVE_load_path} among 200 numerical simulations. These quantities are plotted against the normal displacement jump $\delta_n$.}
	\label{fig:DEMRVE_microinfo}
\end{figure}

The DEM RVE is loaded in the normal $\vec{n}$ and tangential $\vec{m}$ directions of the frictional surface by displacement controls $\delta_n$ and $\delta_m$ (Figure \ref{fig:DEMRVE_load_path_1}). 
The synthetic database consists of 200 numerical experiments under different loading paths. 
They differ from each other in the ratio of normal and tangential loading rate $\dot{\delta_n}/\dot{\delta_m}$, as well as the loading-unloading-reloading cycles, as illustrated in Figure \ref{fig:DEMRVE_load_path_2}, \ref{fig:DEMRVE_load_path_3} and \ref{fig:DEMRVE_load_path_4}. 
The traction-separation curves of the experiments are recored and three examples corresponding to the paths in Figure \ref{fig:DEMRVE_load_path} are presented in Figure \ref{fig:DEMRVE_UT}. 
The microstructural information required for the intermediate nodes in the directed graphs, such as porosity, coordination number and fabric tensor, are also recored during the simulations. 
The open-source library NetworkX \citep{hagberg2008exploring} is employed to analyze the graph of the particle interactions in the RVEs. 
Figure \ref{fig:DEMRVE_microinfo} presents examples of microstructural information and graph characteristics for the three example loading paths. 
The 200 numerical simulations in the database are shuffled. 
The first 50 simulations are used as "training data" for the calibration of model parameters for the edges in the directed graphs. 
The other 150 simulations are "test data" only for evaluating the blind prediction accuracy of the resultant constitutive model. 

\bibliographystyle{plainnat}
\bibliography{main}

\end{document}